\pdfoutput=1

\documentclass[11pt]{article}

\usepackage{ACL2023}

\newcommand{\datasetName}{{\sc InfoSync}\xspace}
\newcommand{\paperTitle}{\datasetName: Information Synchronization across Multilingual Semi-structured Tables}
\usepackage{times}
\usepackage{latexsym}

\usepackage{microtype}
\usepackage{amsmath}
\usepackage{graphicx}

\usepackage{subcaption}
\usepackage{url}
\usepackage{amsmath, amsthm, amssymb}

\usepackage{paralist}

\usepackage{enumitem}
\usepackage{xspace}

\usepackage{paralist}
\usepackage{booktabs} 

\usepackage{wasysym} 
\usepackage{array}

\usepackage{lingmacros}
\newcommand*{\Comb}[2]{{}^{#1}C_{#2}}%
\usepackage{extarrows} 

\usepackage{times}
\usepackage{latexsym}

\usepackage[T1]{fontenc}

\usepackage[utf8]{inputenc}

\usepackage{microtype}

\usepackage{inconsolata}

\title{\paperTitle}

\author{
Siddharth Khincha\textsuperscript{\rm 1},
Chelsi Jain\textsuperscript{\rm 2},
Vivek Gupta\textsuperscript{\rm 3†}\thanks{~~Corresponding Author},
Tushar Kataria\textsuperscript{\rm 3†},
Shuo Zhang\textsuperscript{\rm 4}\\ 
\textsuperscript{\rm 1}IIT Guwahati,
\textsuperscript{\rm 2}CTAE, Udaipur,
\textsuperscript{\rm 3}University of Utah\thanks{~~Equal Contribution},
\textsuperscript{\rm 4}Bloomberg,\\
s.khincha@iitg.ac.in, chelsiworld@gmail.com\\\{vgupta, tkataria\}@cs.utah.edu, \{szhang611\}@bloomberg.net \\
}

\begin{document}
\maketitle
\begin{abstract}
Information Synchronization of semi-structured data across languages is challenging.
For instance, Wikipedia tables in one language should be synchronized across languages. 
To address this problem, we introduce a new dataset \datasetName and a two-step method for tabular synchronization. \datasetName contains 100K entity-centric tables (Wikipedia Infoboxes) across 14 languages, of which a subset ($\sim$3.5K pairs) are manually annotated. 
The proposed method includes 1) \textit{Information Alignment} to map rows and 2) \textit{Information Update} for updating missing/outdated information for aligned tables across multilingual tables. 
When evaluated on \datasetName, information alignment achieves an F1 score of 87.91 (en $\leftrightarrow$ non-en). 
To evaluate information updation, we perform human-assisted Wikipedia edits on Infoboxes for 603 table pairs. Our approach obtains an acceptance rate of 77.28$\%$ on Wikipedia, showing the effectiveness of the proposed method.

\end{abstract}

\section{Introduction}
\label{sec:introduction}

\begin{figure}[t]
       \centering
	\includegraphics[scale=0.40]{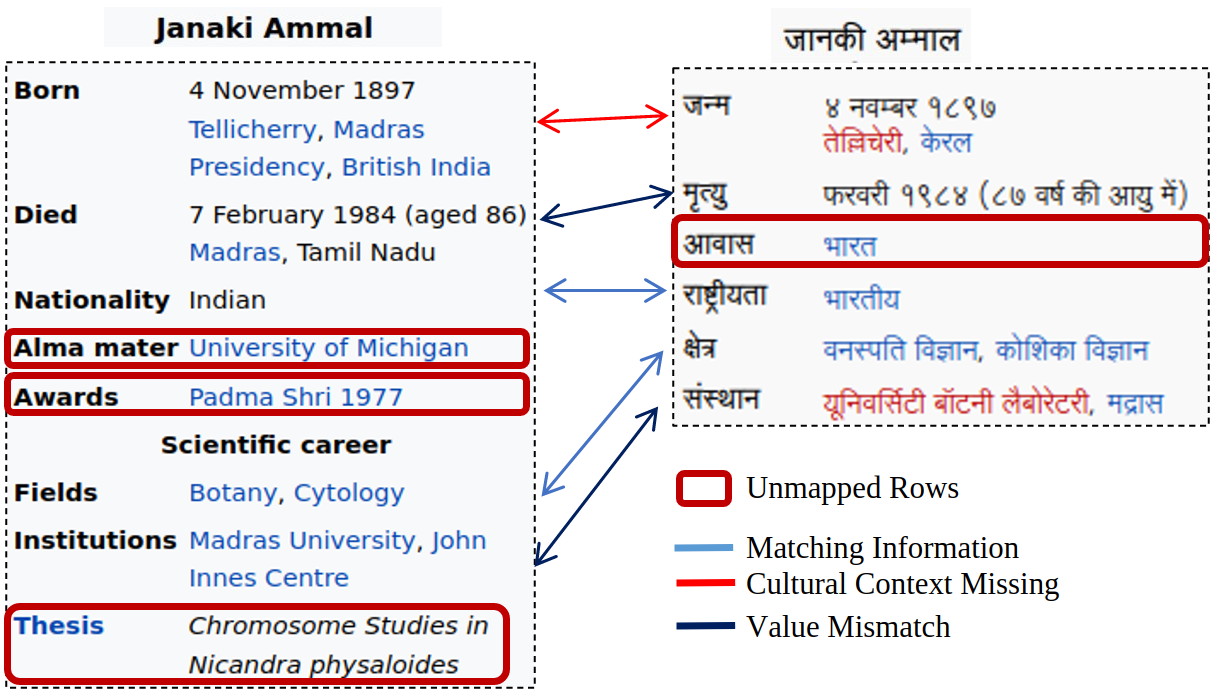}
    \caption{Janaki Ammal Infoboxes in English (right) and Hindi (left). Hindi Table lacks the "British Rule of India" as a cultural context. Two value mismatches (a) The Hindi table doesn't list \textit{Died} key's state (b) Institution values differ. The Hindi table mentions "residence" while the English table doesn't. Hindi Table is missing Thesis, Awards, and Alma Mater keys. Both don't mention parents, early education, or honors.}
\label{intro:infobox}
\vspace{-2em} 
\end{figure}

English articles across the web are more timely updated than other languages on particular subjects. Meanwhile, culture differences, topic preferences, and editing inconsistency lead to information mismatch across multilingual data, e.g., outdated information or missing information \cite{jang2016utilization,nguyen2018automatically}. Online encyclopedia, e.g., Wikipedia, contains millions of articles that need to be updated constantly, involving expanding existing articles, modifying content such as correcting facts in sentences~\citep{Darsh:2019:AFS} and altering Wikipedia categories~\citep{Zhang:2020:GCS}. However, more than 40\% of Wikipedia's active editors are in English. At the same time, only 15\% of the world population speak English as their first language. 
Therefore, information in languages other than English may not be as updated \cite{10.1145/2207676.2208553}. See Figure~\ref{intro:infobox} for an example of an information mismatch for the same entity across different languages. In this work, we look at synchronizing information across multilingual content.

To overcome the above-mentioned problem, we formally introduce the task of Information Synchronization for multilingual articles, which includes paragraphs, tables, lists, categories, and images. But due to its magnitude and complexity, synchronizing all of the information across different modalities on a webpage is daunting. Therefore, this work focuses on semi-structured data, a.k.a. table synchronization in a few languages, as the first step toward our mission. 

We consider Infobox, a particular type of semi-structured Wikipedia tables~\citep{zhang2020web}, which contain entity-centric information, where we observe various information mismatches, e.g.,  missing rows (cf. Figure \ref{intro:infobox}). 
One intuitive idea to address them is translation-based. 
However, the Infoboxes contain rows with implicit context; translating these short phrases is prone to errors and leads to ineffective synchronization \cite{minhas-etal-2022-xinfotabs}.
To systematically assess the challenge, we curate a dataset, namely \datasetName, consisting of 100K multilingual Infobox tables across 14 languages and covering 21 Wikipedia categories. 
$\sim$3.5K table pairs of English to non-English or non-English to non-English are sampled and manually synchronized.

We propose a table synchronization approach that comprises two steps: \begin{inparaenum}[(1.)] \item \textbf{Information Alignment:} align table rows, and \item \textbf{Information Update:} update missing or outdated rows across language pairs to circumvent the inconsistency. \end{inparaenum}
The \textit{information alignment} component aims to align the rows in multilingual tables. 
The proposed method uses corpus statistics across Wikipedia, such as key and value-based similarities.

The \textit{information update} step relies on an effective rule-based approach. We manually curate nine rules: row transfer, time-based, value trends, multi-key matching, append value, high to low resource, number of row differences, and rare keys.
Both tasks are evaluated on \datasetName to demonstrate their effectiveness. Apart from the automatic evaluation, we deploy an online experiment that submits the detected mismatches by our method to Wikipedia after strictly following Wikipedia editing guidelines. We monitor the number of accepted and rejected edits by Wikipedia editors to demonstrate its efficacy. All proposed edits are performed manually, in accordance with Wikipedia's editing policies and guidelines\footnote{\url{https://en.wikipedia.org/wiki/Wikipedia:List_of_policies_and_guidelines}}, rule set\footnote{\url{https://en.wikipedia.org/wiki/Wikipedia:Simplified_ruleset}}, and policies\footnote{\url{https://en.m.wikipedia.org/wiki/Wikipedia:Editing_policy}}. These changes were subsequently accepted by Wikipedia editors, demonstrating the efficacy of our methodology.

The contributions in this work are as follows: 1) We investigate the problem of Information Synchronization across multilingual semi-structured data, i.e., tables, and construct a large-scale dataset \datasetName; 2) We propose a two-step approach (alignment and updation) and demonstrate superiority over exiting baselines; 3) The rule-based updation system achieves excellent acceptance when utilized for human-assisted Wikipedia editing.
Our \datasetName dataset and method source code are available at \url{https://info-sync.github.io/info-sync/}.
\section{Motivation}

\subsection{Challenges in Table Synchronization}
\label{sec:motivation}
We observe the following challenges when taking Wikipedia Infoboxes as a running example. Note this is not an exhaustive list.

    \emph{\textbf{MI: Missing Information}} represents the problem where information appears in one language and is missing in others. This may be due to the fact that the table is out-of-date or to cultural, social, or demographic preferences for modification (cf. Figure~\ref{intro:infobox}).

    \emph{\textbf{OI: Outdated Information}} denotes that information is updated in one language but not others. 

    \emph{\textbf{IR: Information Representation}} varies across languages. For example, one attribute about "parents" can be put in a single row or separate rows ("Father" and "Mother").
    
    \emph{\textbf{UI: Unnormalized Information}} presents cases where table attributes can be expressed differently. For example, "known for" and "major achievements" of a person represent the same attribute (i.e., paraphrase). 
    
    \emph{\textbf{LV: Language Variation}} means that information is expressed in different variants across languages. 
    This problem is further exaggerated by the implicit context in tables when translating. E.g., "Died" in English might be translated to "Overleden" (Pass Away) or "overlijdensplaats" (Place of Death) in Dutch due to missing context.
    
    \emph{\textbf{SV: Schema Variation}} denotes that the schema (template structure) varies. 
    For example, extraction of "awards" in Musician tables can be harrowing due to dynamic on-click lists (\emph{Full Award Lists}). 
    
    \emph{\textbf{EEL: Erroneous Entity Linking}} is caused by mismatched linkages between table entities among multiple languages, e.g., "ABV" and "Alcohol by Volume".

\subsection{Wikipedian "Biases"}
\label{sec:motivation_syncronization}
Wikipedia is a global resource across over 300 languages. However, the information is skewed toward English-speaking countries \cite{roy-etal-2020-topic} as English has the most significant Wikipedia covering 23\% (11\%) of total pages (articles). Most users' edits (76\%) are also done in English Wikipedia. English Wikipedia also has the highest number of page reads (49\%) and page edits (34\%), followed by German (20\% and 12\%) and Spanish (12\% and 6\%), respectively. 
Except for the top 25 languages, the total number of active editors, pages, and edits is less than 1\% \cite{10.1145/2462932.2462959,ALONSO2016114}.

Multilingual Wikipedia articles evolve separately due to cultural and geographical bias \cite{10.1002/asi.21577,reagle2011gender,10.1145/2567948.2576931}, which prevents information synchronization. 
For example, information on "Narendra Modi" (India's Prime Minister) is more likely to be better reflected in Hindi Wikipedia than in other Wikipedias. This means that in addition to the obvious fact that smaller Wikipedias can be expanded by incorporating content from larger Wikipedias, larger Wikipedias can also be augmented by incorporating information from smaller Wikipedias.
Thus, information synchronization could assist Wikipedia communities by ensuring that information is consistent and of good quality across all language versions.
\section{The \datasetName Dataset}
\label{sec:dataset}
To systematically assess the challenge of information synchronization and evaluate the methodologies, we aim to build a large-scale table synchronization dataset \datasetName based on entity-centric Wikipedia Infoboxes.

\subsection{Table Extraction}
We extract Wikipedia Infoboxes from pages appearing in multiple languages on the same date to simultaneously preserve Wikipedia's original information and potential discrepancies. These extracted tables are across 14 languages and cover 21 Wikipedia categories.  

\textbf{Languages Selection.} We consider the following languages English(en), French(fr), German(de), Korean(ko), Russian(ru), Arabic(ar), Chinese(zh), Hindi(hi), Cebuano(ceb), Spanish(es), Swedish(sv), Dutch(nl), Turkish(tr), and Afrikaans(ak). We extracted tables across 14 languages and covered 21 diverse Wikipedia categories. In these 14 languages, four are low resource (af, ceb, hi, tr)  < 6000, seven of them medium resource (ar, ko,nl, sv, zh,ru, de,es) (6000–10000), and the remaining one are high resource (en, en, fr), w.r.t. to the number of infobox total tables (see Table 1 in paper). Our choices were motivated by the following factors:- a) Cover all the continents, thus covering the majority and diverse population. Out of chosen languages, 7 (English, French, German, Spanish, Swedish, Dutch, and Turkish) are European. b). They have sufficient pages with info boxes; each entity info box is present in at least five languages, and c) an adequate number of rows (5 and above) facilitates better data extraction. 

\textbf{Categories.} Extracted tables cover twenty-one simple, diverse, and popular topics: \textit{Airport, Album, Animal, Athlete, Book, City, College, Company, Country, Food, Monument, Movie, Musician, Nobel, Painting, Person, Planet, Shows, and Stadiums}. We observe that \textit{Airport} has the most number of entity tables followed by \textit{Movie} and \textit{Shows}, as shown in Table~\ref{tab:apdx_infosync_category_wise_entities}. Other extraction details are provided in Appendix~\ref{apdx:ted}.

\subsection{Tabular Information Mismatched} 
\label{ssec:tablular_information_mismatched}

\begin{table}[h]
\centering
\small
\setlength{\tabcolsep}{3.5pt}
\begin{tabular}{ l|cc | cc} 
\toprule
\bf Ln & \multicolumn{2}{c}{\bf Average Table Transfer \%} & \multicolumn{2}{|c}{\bf Language Statistics} \\
\bf C1     & \bf C1 $\rightarrow$ $\sum_L ln$  & \bf $\sum_L ln$   $\rightarrow$ C1  & \bf $\#$ Tables & \bf AR   \\
\midrule
 af             & 17.46  & 400.5  & 1575   & 9.91        \\ 
 ar             & 34.02  & 27.38  & 7648   & 13.01       \\ 
 ceb            & 42.87  & 134.88  & 3870   & 7.82      \\ 
 de             & 40.73  & 27.12  & 8215   & 7.88       \\ 
 en             & 45.85  & 0.32  & 12431  & 12.60        \\
 es             & 38.78  & 9.00  & 9920   & 12.59            \\ 
 fr             & 41.25  & 4.73 & 10858  & 10.30          \\ 
 hi             & 18.39  & 358.97  & 1724   & 10.91      \\ 
 ko             & 31.13  & 40.51 & 6601   & 9.35          \\ 
 nl             & 33.69  & 24.6  & 7837   & 10.46        \\ 
 ru             & 36.98  & 14.54 & 9066   & 11.41         \\ 
 sv             & 35.53  & 24.62  & 7985   & 9.89       \\ 
 tr             & 28.99  & 59.33   & 5599  & 10.14      \\ 
 zh             & 32.16  & 32.71  & 7140  & 12.43       \\ 
\bottomrule
\end{tabular}
\vspace{-0.5em}
\caption{ \textbf{Average Table Transfer}:- Column 2 shows the average number of tables missing in other languages which can be transferred from C1. Column 3 shows the average number of tables missing in C1, which we can transfer from all languages to C1. Here $L$ is the set of all languages ($ln$) except source or transfer language. \textbf{Language Statistics}:- The number of tables and average rows (AR) per table across different categories for each language.}
\label{tab:infosync_table_transfer_and_language_wise}
\vspace{-1.0em}
\end{table}

We analyze the extracted tables in the context of the synchronization problem and identify the information gap. 
The number of tables is biased across languages, as shown in Table \ref{tab:infosync_table_transfer_and_language_wise}. We observe Afrikaans, Hindi, and Cebuano have a significantly less number of tables.
Similarly, the table size is biased across several languages. Dutch and Cebuano have the last rows.
In addition, the number of tables across categories is uneven; refer to Table \ref{tab:apdx_infosync_category_wise_Tables}. Airport and Movie have the highest number of tables. Table \ref{tab:apdx_infosync_category_wise_Tables} also reports the average number of rows for a category. Planet, Company, and Movie have the highest average number of rows. 

\begin{table}[!h]
\setlength{\tabcolsep}{2.5pt}
\centering

\small
\begin{tabular}{ l|cc|l|cc } 
\toprule
\bf Topic    & \bf $\#$ Tables   &  \bf AR & \bf Topic    & \bf $\#$ Tables   &  \bf AR  \\ 
\midrule
 Airport             & 18512      & 9.66  &  Diseases            & 3973       & 6.03  \\
Food                & 6184       & 7.93  & Monument            & 1550       & 9.71 \\
 Album               & 5833       & 7.58 & Medicine            & 2516       & 15.20    \\ 
 Animal              & 3304       & 8.27 &  Movie               & 12082      & 13.29  \\ 
 Athlete             & 3209       & 9.09  &  Musician            & 2729       & 9.53  \\ 
 Book                & 1550       & 9.99  &  Nobel               & 9522       & 9.84  \\
 Painting            & 3542       & 7.05   & Country             & 3338       & 22.85  \\ 
 City                & 3088       & 14.45 & Person              & 2252       & 11.87  \\ 
 College             & 1857       & 11.01  & Planet              & 1233       & 16.80  \\ 
 Company             & 2225       & 13.85  & Shows               & 5644       & 13.86 \\
 Stadium & 6326 &  10.94 & & & \\ 

\hline
\end{tabular}
\vspace{-0.5em}
\caption{\textbf{Category Statistics} :- Number of tables in each category and average number of rows (AR) across different languages.}
\label{tab:apdx_infosync_category_wise_Tables}
\vspace{-1.5em}
\end{table}

When synchronizing a table from one language to another, we observe that the maximum number of tables can be transferred from English, French, and Spanish from Column 1 in Table~\ref{tab:infosync_table_transfer_and_language_wise}. Afrikaans, Hindi, and Cebuano have the least overlapping information (Column 3) with all other languages. 
The number of rows (Column 5) varies substantially between languages, with Spanish and Arabic having the highest number.

\subsection{\datasetName Evaluation Benchmark} 

We construct the evaluation benchmark by manually mapping the table's pairs in two languages. The table pairs we consider can be broadly split into English $\leftrightarrow$ Non-English and Non-English $\leftrightarrow$ Non-English. The annotations are conducted as follows. 

\paragraph{English $\leftrightarrow$ Non-English:} We sample 1964 table pairs, where a minimum of 50 pairs for each category and language are guaranteed. We divide the annotated dataset, ratio of $1:2$, into validation and test sets. The non-English tables are translated into English first and then compared against the English version. Furthermore, native speakers annotated 200 table pairs for English $\xleftrightarrow{*}$ Hindi and English $\xleftrightarrow{*}$ Chinese to avoid minor machine translation errors. 
          
\paragraph{Non-English $\leftrightarrow$ Non-English:} We consider six non-English languages: two from each High resource (French, Russian), Medium Resource (German, Korean), and Low Resource (Hindi, Arabic), w.r.t. the number of tables in \datasetName. We sample and annotate 1589 table pairs distributed equally among these languages, where we choose an average of $\sim 50$ tables for all pairs of languages. Both are translated into English before manually mapping them.

In addition, for more detailed analysis, we also annotate metadata around table synchronization challenges such as MI, IR, LV, OI, UI, SV, and EEL, as discussed in \S \ref{sec:motivation}. 
\section{Table Synchronization Method}
\label{sec:method_details}

This section will explain our proposed table synchronization method for addressing missing or outdated information. 
This method includes two steps: information alignment and update. The former approach aims to align rows across a pair of tables, and the latter helps to update missing or outdated information. We further deploy our update process in a human-assisted Wikipedia edit framework to test the efficacy in the real world.


\subsection{Information Alignment}
\label{ssec:method_alignment}

An Infobox consists of multiple rows where each row has a key and value pair.
Given a pair of tables $T_x = [..., (k_x^i, v_x^i), ...]$ and $T_y = [..., (k_y^j, v_y^j), ...]$ in two languages, table alignment aims to align all the possible pairs of rows, e.g., $(k_x^i, v_x^i)$ and $(k_y^j, v_y^j)$ refer to the same information and should be aligned. 

We propose a method that consists of five modules, each of which relaxes matching requirements in order to create additional alignments.

{\bf \emph{M1. Corpus-based.}} 
The pair of rows $(k_x, v_y)$ in $T_x$ and $(k_y, v_y)$ in $T_y$ are supposed to be aligned if $cosine(em({tr_x^{en}(k_x})), em(tr_y^{en}({k_y}))) > \theta_1$, where $em$ is the embedding, $\theta_1$ is the threshold, and $tr_y^{en}()$ denotes the English translation of $k$ if $k$ is not in English.
In order to achieve accurate key translations, we adopt a majority voting approach, considering multiple translations of the same key from different category tables.
We consider the key's values and categories as additional context for better translation during the voting process. To simplify the voting procedure, we pre-compute mappings by selecting only the most frequent keys for each category across all languages.

{\bf \emph{M2. Key-only.}}
This module attempts to align the unaligned pairs in module 1. Using their English translation, it first computes cosine similarity for all possible key pairs. 
$k^x$ will be aligned to $k^y$ only if they are mutually most similar key and the similarity is above a certain threshold $\theta_2$.

This is similar to maximum bipartite matching, treating similarity scores as edge weights followed by threshold-based pruning. And it ensures we are capturing the highest similarity mapping from both language directions. Note that here we use only keys as the text for similarity computation.

{\bf \emph{M3. Key value bidirectional.}} This module is similar to step 2, except it uses the entire table row for computing similarities, i.e., key + value, using threshold $\theta_3$. 

{\bf \emph{M4. Key value unidirectional.}}  
This module further relaxes the bidirectional mapping constraint in step 3, i.e., thus removing the requirement of the highest similarity score matching from both sides.
We shift to unidirectional matching between row pairs, i.e., consider the highest similarity in either direction. However, this may result in adding spurious alignments. To avoid this, we have a higher threshold ($\theta_4$) than the prior step.

{\bf \emph{M5. Multi-key.}}
Previous modules only take the most similar key for alignment if exceeding the threshold. In this module, we further relax the constraint to select multiple keys (maximum two), given exceeding a threshold ($\theta_5$). Multi-key mapping is sparse, but the above procedure will lead to dense mapping. To avoid this, we introduce a \textit{soft constraint} for value-combination alignment, where multi-key values are merged. We consider valid multi-key alignment when the merge value-combination similarity score exceeds that of the most similar key. 

The thresholds of five modules are tuned in the sequence as stated above.

\begin{figure*}
    \centering
    \includegraphics[scale=0.39]{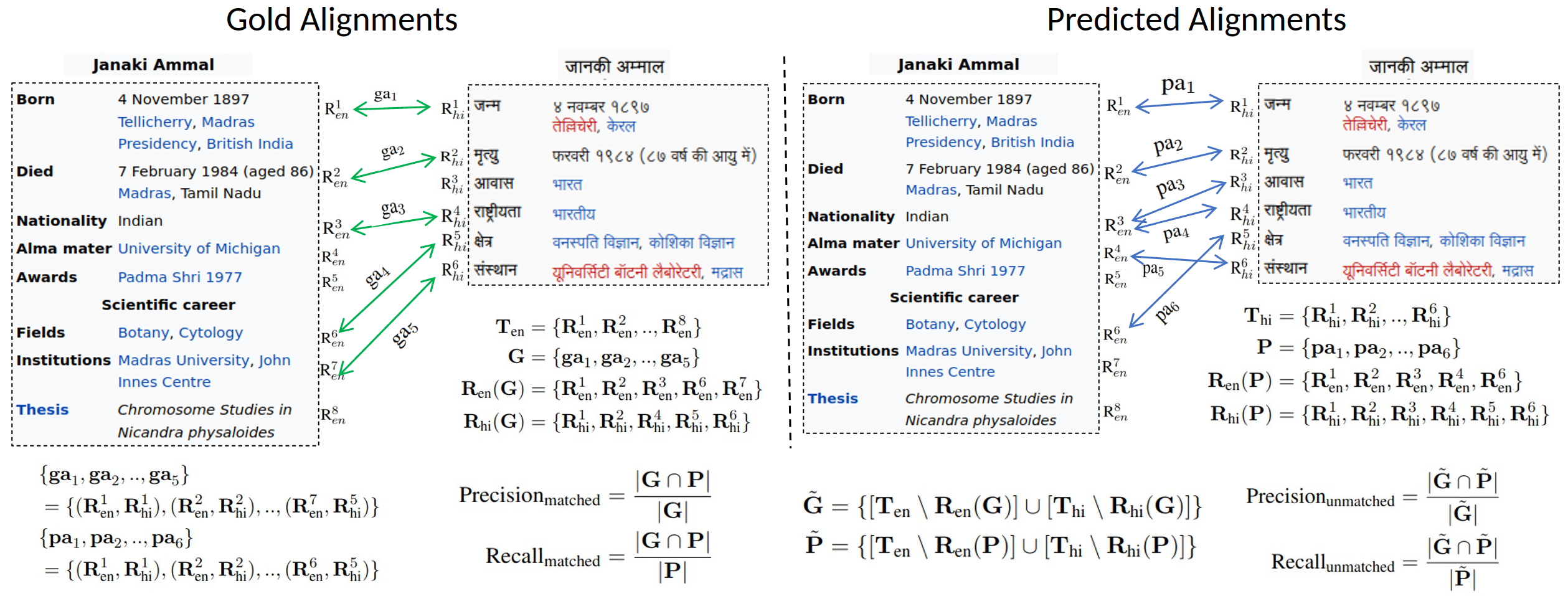}
    \vspace{-0.5em}
    \caption{Explanation of Alignment Performance Metrics:-  $\mathbf{T}_{en}$ and $\mathbf{T}_{hi}$ is a set of all rows in English and Hindi Table, respectively. $\mathbf{R}_{x}^n$ represent $n^{th}$ row in the table of language $x$. $\mathbf{R}_{x}(\mathbf{X})$ returns all the rows in language $x$ using mapping $\mathbf{X}$. $|.|$ represents the cardinality of the set. $\mathbf{G}$ is a set of all manual alignments. Each alignment is stored as a tuple ${\mathbf{ga}}_i$ = $(\mathbf{R}_{x}^m, \mathbf{R}_{x}^n)$. $\mathbf{P}$ is a set of predicted alignments (can see there are mistakes in the alignment).}
    \label{fig:task}
    \vspace{-1.0em}
\end{figure*}

\subsection{Information Updation}
\label{ssec:method_updation}
Information modification includes \textit{Row Append} (adding missing rows), \textit{Row Update} (replacing or adding values), and \textit{Merge} Rows.
We propose a rule-based heuristic approach for information updates. The rules are in form of logical expression ($\forall_{(\text{R}_{T_x},\text{R}_{T_y})}$ L $\mapsto$ R) applied on infobox tables, where, R$_{T_x}$ and R$_{T_y}$ represent table rows for language $x$ and $y$ respectively. These rules are applied sequentially according to their priority rank (P.R.). Rules explanations are described below.

\emph{\textbf{R1. Row Transfer.}} Following the logistic rule of 
\begin{align*}
   \forall_{(\text{R}_{T_x},\text{R}_{T_y})} \text{Al}^{T_y}_{T_x}(\text{R}_{T_x};\text{R}_{T_y}) = 0 \\ \mapsto {T_y \cup tr{_{x}^{y}}(\text{R}_{T_x})} \bigwedge \text{Al}^{T_y}_{T_x}(\text{R}_{T_x}; tr{_{x}^{y}}(\text{R}_{T_x}))=1 
\end{align*}
\noindent , where $\text{Al}^{T_y}_{T_x}(.;.)$ represents the alignment mapping between two tables ${T_y}$ and ${T_x}$. Unaligned rows are transferred from one table to another. 

\emph{\textbf{R2. Multi-Match.}}

We update the table by removing multi-alignments and replacing them with merged information to handle multikey alignments.

\emph{\textbf{R3. Time-based.}} We update aligned values using the latest timestamp.

\emph{\textbf{R4. Trends (positive/negative).}} This update applies to cases where the value is highly likely to follow a monotonic pattern (increasing or decreasing) w.r.t. time, e.g., athlete career statistics. The authors curated the positive/negative trend lists.

\emph{\textbf{R5. Append Values.}} Additional value information from an up-to-date row is appended to the outdated row.

\emph{\textbf{R6. HR to LR.}} This rule transfers information from high to low resource language to update outdated information.

\emph{\textbf{R7. $\#$Rows.}} This rule transfers information from bigger (more rows) to smaller (fewer rows) tables.  

\emph{\textbf{R8. Rare Keys (Non Popular).}} 
We update information from the table where non-popular keys are likely to be added recently to the outdated table. The authors also curate non-popular keys.

Detailed formulation of logical rules and their priority ranking are listed in Table \ref{tab:apdx_updation_rules}. Figure \ref{figure::apdx:Update} in Appendix shows an example of table update. 

\begin{table*}[h]
\small
\setlength{\tabcolsep}{2.5pt}
\begin{tabular}{ clcr } 
\toprule
\bf P.R.              & \bf Rule Name    & \bf Logical Rule $\forall_{(\text{R}_{T_x},\text{R}_{T_y})}$ L $\mapsto$ R    & \bf    Update Type \\
\midrule

  1        & Row Transfer & $\forall_{(\text{R}_{T_x},\text{R}_{T_y})} \text{Al}^{T_y}_{T_x}(\text{R}_{T_x};\text{R}_{T_y}) = 0 $ & Row Addition   \\ 
 & & $\mapsto {T_y \cup tr{_{x}^{y}}(\text{R}_{T_x})}$ $\bigwedge$ $\text{Al}^{T_y}_{T_x}(\text{R}_{T_x}; tr{_{x}^{y}}(\text{R}_{T_x}))=1$ &  \\\midrule

 2        &  Multi-Match & $\forall_{(\text{R}_{T_x},\text{R}_{T_y})} ( \sum_{\text{R}_{T_y}} \text{Al}^{T_y}_{T_x}(\text{R}_{T_x};\text{R}_{T_y}) ) > 1$ & Row Delete   \\ 
 & & $\mapsto \{T_y\setminus\cup_{( \forall_{\text{R}_{T_y}} \text{Al}^{T_y}_{T_x}(\text{R}_{T_x};\text{R}_{T_y}) = 1 )}\text{R}_{T_y} \}\bigcup  tr{_{x}^{y}}(\text{R}_{T_x}) \bigwedge \text{Al}^{T_y}_{T_x}(\text{R}_{T_x}; tr{_{x}^{y}}(\text{R}_{T_x}))=1$ & \\ 
 
3      &  Time-based & $\forall_{(\text{R}_{T_x}, \text{R}_{T_y})} \text{Al}^{T_y}_{T_x}(\text{R}_{T_x};\text{R}_{T_y}) = 1 \bigwedge (\text{isTime}(\text{R}_{T_x},\text{R}_{T_y})=1)$ & Value Substitute    \\ 
 & & $  \bigwedge (\text{exTime}(\text{R}_{T_x}) > \text{exTime}(\text{R}_{T_y}))  \mapsto   \text{R}_{T_y} \leftarrow  tr{_{x}^{y}}(\text{R}_{T_x})$ & \\

4 & Positive Trend & $\forall_{(\text{R}_{T_x},\text{R}_{T_y}, \text{PosTrend})} \text{Al}^{T_y}_{T_x}(\text{R}_{T_x};\text{R}_{T_y})  = 1 \bigwedge \text{exKey}(\text{R}_{T_x}) \in \text{PosTrend}$  & Value Substitute \\ 
& or & $ \bigwedge \text{R}_{T_x} > \text{R}_{T_y} \mapsto \text{R}_{T_y} \leftarrow \text{R}_{T_x} $ & \\

 & Negative Trend & $\forall_{(\text{R}_{T_x},\text{R}_{T_y}, \text{NegTrend})} \text{Al}^{T_y}_{T_x}(\text{R}_{T_x};\text{R}_{T_y})  = 1 \bigwedge \text{exKey}(\text{R}_{T_x}) \in \text{NegTrend}$  & Value Substitute \\ 
& & $ \bigwedge \text{R}_{T_x} < \text{R}_{T_y}  \mapsto \text{R}_{T_y} \leftarrow \text{R}_{T_x} $ & \\
 5        &  Append Value & $\text{R}_{T_x} =\text{V} \bigwedge \forall_{(\text{R}_{T_x}, \text{R}_{T_y})} \text{Al}^{T_y}_{T_x}(\text{R}_{T_x};\text{R}_{T_y}) = 1 \bigwedge |\text{R}_{T_x}[k]| > |\text{R}_{T_y}[k]|$ & Value Addition   \\ 
 & & $ \mapsto  \forall_{(v \in \text{R}_{T_x}[k] ~ \wedge ~ \notin tr{_{x}^{y}}(\text{R}_{T_x}[k]))} \text{R}_{T_y} \leftarrow  \text{R}_{T_y} \cup tr{_{x}^{y}}(v)$ & \\

6        & HR to LR & $({T_x}, {T_y}) \in (HR, LR) \bigwedge \forall_{(\text{R}_{T_x}, \text{R}_{T_y})} \text{Al}^{T_y}_{T_x}(\text{R}_{T_x};\text{R}_{T_y}) = 1  $ & Value Substitute   \\ 
 & & $ \bigwedge tr{_{x}^{en}}(\text{R}_{T_x}) \neq tr{_{y}^{en}}(\text{R}_{T_y}) \mapsto  \text{R}_{T_y} \leftarrow tr{_{x}^{y}}(\text{R}_{T_x})$ & \\

 7        &  $\#$ \text{Rows} & $|T_x| >> |T_y| \bigwedge \forall_{(\text{R}_{T_x}, \text{R}_{T_y})} \text{Al}^{T_y}_{T_x}(\text{R}_{T_x};\text{R}_{T_y}) = 1 \bigwedge tr{_{x}^{en}}(\text{R}_{T_x}) \neq tr{_{y}^{en}}(\text{R}_{T_y}) $ & Value Substitute  \\ 
& & $ \mapsto  \text{R}_{T_y} \leftarrow tr{_{x}^{y}}(\text{R}_{T_x})$ & \\

 8 & Rare Keys & 
$\forall_{(\text{R}_{T_x},\text{R}_{T_y}, \text{RarKeys})} \text{Al}^{T_y}_{T_x}(\text{R}_{T_x};\text{R}_{T_y})  = 1 \bigwedge \text{tr}^{en}_x(\text{R}_{t_x})\neq \text{tr}^{en}_y(\text{R}_{t_y})$ & Value Substitute \\
&& $ \bigwedge \forall_{(\text{R}_{T_x},\text{R}_{T_y}} |\text{exKey}(\text{R}_{T_x}) \in \text{RarKey}| > |\text{exKey}(\text{R}_{T_y}) \in \text{RarKey}| \mapsto \text{R}_{T_y} \leftarrow \text{R}_{T_x}$ &\\
\bottomrule
\end{tabular}
\caption{\textbf{Logical Rules for Information Updation}. Notation:- $T_z$ represents a table in language $z$, $\text{R}_{T_z}$} represents a row of the table. In $\text{R}_{T_z}[k]=v$, $k$,$v$ represent key and value pair. For $\text{R}_{T_z}[k]=V$, $V$ denotes value list mapped to a key $k$. $\text{Al}^{T_y}_{T_x}(.;.)$ represents the alignment mapping between two tables ${T_y}$ and ${T_x}$. Translation between two languages($p$ and $q$) is represented by $tr^p_q(.)$. \emph{exKey} extract key from a table row. \emph{isTime} is true if the row has time entry. \emph{exTime} extract time from table row. \emph{PosTrend/NegTrend} represent list of keys whose value always increase or decrease with time. \emph{RarKey} represent set of keys are least frequent in the corpora.
\label{tab:apdx_updation_rules}
\vspace{-0.5em}
\end{table*}

\paragraph{Human-assisted Wikipedia Infobox Edits:} We apply the above rules to assist humans in updating Wikipedia infoboxes. Following Wikipedia edit guidelines\footnote{\url{https://en.wikipedia.org/wiki/Wikipedia:List_of_policies_and_guidelines}}, rule set\footnote{\url{https://en.wikipedia.org/wiki/Wikipedia:Simplified_ruleset}}, and policies\footnote{\url{https://en.m.wikipedia.org/wiki/Wikipedia:Editing_policy}}, we append our update request with a description to provide evidence, which contains
\begin{inparaenum}[(a)]
\item up-to-date entity page URL in the source language,
\item exact table rows information, the source language, and the details of the changes,
\item and one additional citation discovered by the editor for extra validation. \footnote{We use a search engine such as Google, Bing, you.com, perplexity.ai, find, etc. for additional citation.}
\end{inparaenum} We further update beyond our heuristic-based rules but are aligned through our information alignment method. 

\section{Experiments} 
\label{sec:exp}

Our experiments assess the efficacy of our proposed two-stage approach by investigating the following questions.

- What is the efficacy of the unsupervised multilingual method for table alignment? (\S\ref{ssec:exp_alignment})

- How significant are the different modules of the alignment algorithm? (\S\ref{ssec:exp_alignment} and \S\ref{apx:ablation_study})

- Does the rule-based updating approach effective for information synchronization? (\S\ref{ssec:exp_updation})

- Can the two-step approach assist humans in updating Wikipedia Infoboxes? (\S\ref{ssec:exp_updation})

\begin{table*}[!h]
\small
\centering
\setlength{\tabcolsep}{1.5pt}
\begin{tabular}{ l|cccc | cccc } 
\toprule
 \bf Method & & \multicolumn{2}{c}{\bf Match} & &  &\multicolumn{2}{c}{\bf UnMatch} & \\
 
 & \bf  $T_{en}$ $\leftrightarrow$ $T_{x}$ & \bf  $T_{x}$ $\leftrightarrow$ $T_{y}$ & \bf $T_{en}$ $\xleftrightarrow{*}$ $T_{hi}$ & \bf  $T_{en}$ $\xleftrightarrow{*}$ $T_{zh}$ & \bf  $T_{en}$ $\leftrightarrow$ $T_{x}$ & \bf  $T_{x}$ $\leftrightarrow$ $T_{y}$ & \bf  $T_{en}$ $\xleftrightarrow{*}$ $T_{hi}$ & \bf  $T_{en}$ $\xleftrightarrow{*}$ $T_{zh}$ \\
\hline
SimCSE & 75.78 & 68.46 & 77.93 & 80.47  & 79.11 & 76.3 & 73.31 &74.91\\
LaBSE    & 85.25 & 78.44 & 88.98 & 89.1 & 87.03 & 81.7 & 88.98 & 85.06\\

mBERT-mp  & 80.98 & 73.74 & 82.9 & 86.73 & 82.68 & 80.22 & 76.73 & 81.85\\
XLM-R & 83.38 & 75.02 & 86.85 & 88.08 & 85.42 & 80.65 & 83.14 & 83.1\\
MPNet & 82.85 & 78.63 & 86.08 & 87.58 & 84.2 & 83.45 & 83.14 & 83.76\\
distill mBERT & 84.55 & 77.45 & 87.64 & 88.7 & 86.3 & 82.28 & 83.14 & 84.3\\
\hline
\bf Our Approach\\
 Corpus-based  & 61.86 & 56.74 & 57.34 & 69.33 & 70.51 & 71.73 & 54.01 & 63.11 \\ \hline
+ Key Only   & 70.41	 & 62.14 & 73.4  & 74.67 & 73.85 & 73.52 & 62.49 & 66.23 \\
+ Key-Val-Bi & 87.71 & 84.2  & 90.07 & 93.04 & 89.51 & 85.52 & 85.06 & 89.2\\
+ Key-Val-Uni& 87.89 & 84.33 & 90.34 & 93.12 & 89.52 & 85.42 & 85.16 & 88.62  \\
+ Multi-Key	 & \bf87.91 & \bf 84.36 & \bf 90.14 & \bf 92.8  & \bf 89.3  & \bf 85.46 & \bf 84.98 & \bf 88.15 \\
\bottomrule
\end{tabular}
\vspace{-0.5em}
\caption{\textbf{Matched and UnMatch Score :} F1-Score for all test sets of \datasetName.}
\label{tab:aligned_unaligned_results}
\vspace{-2.0em}
\end{table*}

\begin{table*}[h]
\small
\centering
\setlength{\tabcolsep}{6.5pt}
\begin{tabular}{ l|cccccc|ccccccc } 
\bottomrule
\bf Method & &\multicolumn{4}{c}{\bf $T_{en}$ $\leftrightarrow$ $T_{x}$ }& & & \multicolumn{4}{c}{\bf  $T_{x}$ $\leftrightarrow$ $T_{y}$}  &\\
& \bf OI    & \bf IR    &  \bf SV & \bf LV     & \bf UI    & \bf EL  & \bf \bf OI    & \bf IR    &  \bf SV & \bf LV     & \bf UI    & \bf EL \\
\midrule
 w/o Align         & 298   & 286   & 22    & 158       & 388   & 118 & 245   & 226   & 33    & 146    & 486   & 148 \\ \hline
 Corpus-based              & 81   & 284   & 15    & 141    & 337   & 74  & 108   & 218   & 26    & 102    & 366   & 109 \\ 
 +Key Only               & 110   & 281   & 7     & 120    & 262   & 48  & 77    & 212   & 19    & 94     & 284   & 97 \\
 +Key-Val-Bi & 75    & 232.33 & 6     & 35     & 108   & 8  & 44    & 197   & 15    & 28     & 60    & 18  \\
 +Key-Val-Uni & 74    & 206.67 & 6     & 30     & 99    & 8 & 43    & 188   & 15    & 28     & 59    & 17 \\
 +Multi-Key	            &\bf 74    & \bf179.67 &\bf 6     &\bf 30     &\bf 99    &\bf 8  &\bf 43    &\bf 180.33 &\bf 15    &\bf 28     &\bf 59    &\bf 17  \\ 
\bottomrule
\end{tabular}
\vspace{-0.5em}
\caption{\textbf{Error Analysis for Matched Score :} $T_{en}$ $\leftrightarrow$ $T_{x}$  and $T_{x}$ $\leftrightarrow$ $T_{y}$.}
\label{tab:aligned_error_anlysis}
\vspace{-1.5em}
\end{table*}

\subsection{Experimental Setup}

\paragraph{Baselines Models.} We compare our approach with LaBSE \cite{feng-etal-2022-language}, and SimCSE \cite{gao-etal-2021-simcse}, multilingual sentence transformers embeddings \cite{reimers-gurevych-2020-making} in which we include mBERT (case2) with mean pooling (mp)  \cite{reimers-2020-multilingual-sentence-bert}, and its distill versions (distill mBERT) \cite{sanh2019DistilBERTAD} all in base form. We also compared with XLM-RoBERTa (XLM-R) \cite{DBLP:journals/corr/abs-1911-02116} with mean pooling, and its distill version \cite{reimers-2019-sentence-bert} trained via MPNet-based teacher model (MPNet) \cite{NEURIPS2020_c3a690be}. For all baseline implementation, we use the Hugging Face transformers \cite{wolf-etal-2020-transformers}  and sentence transformers \cite{reimers-2019-sentence-bert} library for the multilingual models' implementation.

\paragraph{Hyper-parameter Tuning.} For our method, we embed translated English keys and values using MPNet model \cite{reimers-gurevych-2019-sentence}. We tune the threshold hyper-parameters using the validation set, $\frac{1}{3}$ of the total annotated set. We sequentially tune the hyper-parameters thresholds ($\theta_1$ to $\theta_5$) in modules training order. Optimal threshold after tuning are $\theta_1$ = (0.8,0.8); $\theta_2$ = (0.64, 0.6); $\theta_3$= $\theta_3$ = (0.54, 0.54); $\theta_4$ =(0.9, 0.54); $\theta_5$ = (0.88, 0.96) for $\text{T}_{en}$ $\leftarrow$ $\text{T}_{x}$ and $\text{T}_{x}$ $\leftarrow$ $\text{T}_{y}$ respectively. We retain the default setting for other models' specific hyperparameters.

\paragraph{Information Alignment.} We consider English as our reference language for alignment. Specifically, we translate all multilingual tables to English using an effective table translation approach of XInfoTabS  \cite{minhas-etal-2022-xinfotabs}. Then, we apply incremental modules as discussed in \S \ref{ssec:method_alignment}.

We tune independently on the validation set for Non-English $\leftrightarrow$ Non-English and English $\leftrightarrow$ Non-English. 

The method is assessed on two sets of metric \begin{inparaenum}[(a.)] \item matched score: measure the F1-score between ground truth matched row and predicted alignment, and \item unmatched score: measure the F1-score between independent (unmatched) rows in ground truth with predicted unaligned rows. See Figure~\ref{fig:task} for the explanations of these metrics.\end{inparaenum}

\paragraph{Information Updation.}  
We apply the heuristic-based approach and deploy the predicted updates 
for human-assisted edits on Wikipedia Infoboxes. 532 table pairs are edited  distributed among $T_{en}$ $\rightarrow$ $T_{x}$, $T_{x}$ $\rightarrow$ $T_{y}$, and $T_{x}$ $\rightarrow$ $T_{en}$, where $x$ and $y$ are non-English languages. 

\subsection{Information Alignment} 
\label{ssec:exp_alignment}

\begin{table}[!h]
\small
\centering
\begin{tabular}{ l|cc } 
\toprule
   \bf Method       & \bf $T_{en}$ $\leftrightarrow$ $T_{x}$    & \bf $T_{x}$ $\leftrightarrow$ $T_{y}$     \\
\midrule
 Corpus-based              &\bf 157   &\bf 245         \\ 
 +Key Only               & 422    & 343    \\
 +Key-Value-Bi           & 526    & 399      \\
 +Key-Value-Uni          & 572    & 415    \\
 +Multi-Key	            & 619    & 437       \\
\bottomrule
\end{tabular}
\caption{\textbf{Error Analysis for UnMatch Score :} Total Unaligned mistakes for $T_{en}$ $\leftrightarrow$ $T_{x}$  and $T_{x}$ $\leftrightarrow$ $T_{y}$.}
\label{tab:unaligned_error_anlysis}
\vspace{-1.5em}
\end{table}

\noindent
\emph{\textbf{Algorithm Efficacy.}} 
Table \ref{tab:aligned_unaligned_results} reports the matched and unmatched scores. 
For match scores, we observe that the corpus-based module achieves an F1 score exceeding 50 for all language pairs. 
Using a key-only module boosts the performance by about 5-15 points. 
Taking the whole row context (key-value pair) with strict constraints on bidirectional mapping, i.e., two-way similarity, improves performance substantially (more than 16 points). Further relaxing the bi-direction constraint to uni-directional matching (one-way similarity), we improve our results marginally with less than 0.5 performance points. Thus relaxation of the bi-direction mapping constraint doesn't lead to significantly better alignments. The multi-key module, which considers one-to-many alignments, further improves the accuracy marginally. The reason for the marginal improvements is very few instances of one-to-many mappings. 

For unmatch scores, we see similar results to match scores. The only significant difference is in key-only performance, where we observe a 0.5x performance improvement compared to match scores. We also analyze the precision-recall in Tables \ref{tab:appendix_eng_x_alignment}, \ref{tab:appendix_x_y_alignment}, \ref{tab:appendix_en_hi_alignment} and \ref{tab:appendix_en_zh_alignment} of Appendix \S\ref{apdx:prec_recall}. We observe that the precision reduces and recall increases for match scores with module addition, whereas the reverse is true for unmatch scores. The number of alignments increases as we add more modules with relaxed constraints. This increases the number of incorrect alignments reducing the precision but increasing the recall. \footnote{There are more incorrect alignments $\Comb{N}{2}$ compared to correct alignments which is $O(n)$.} Similarly, we can note the accuracy of unaligned rows increases because more incorrect alignments are added with relaxed constraints. We also report each module coverage in Appendix~\ref{apdx:coverage}. The performance of our proposed approach grouped by languages, category, and rows keys are detailed in Appendix~\ref{apdx:domain_language_wise}.

\emph{\textbf{Error Analysis.}}
Error analysis (cf \S \ref{sec:motivation}) for matched and unmatched are reported in Table \ref{tab:aligned_error_anlysis} and \ref{tab:unaligned_error_anlysis}, respectively. 
Our proposed method works sequentially, relaxing constraints, 
and the number of falsely aligned rows increases with module addition (cf. Table \ref{tab:unaligned_error_anlysis}). 
Different modules contribute unequally to unaligned mistakes, (25\%, 56\%) of the mistakes come from corpus-based module, (39\%, 22\%) from Key Only Module, (17\%, 35\%) from Key-Value-Bidirectional module, (7\%, 4\%) from Key-Value-uni-directional module, and (7.6\%,  5\%) from multi-key alignment module, for $T_{en}$ $\leftrightarrow$ $T_{x}$ and $T_{x}$ $\leftrightarrow$ $T_{y}$ respectively. The corpus-based module is worst performing in $T_{x}$ $\leftrightarrow$ $T_{y}$ because of difficulty in multilingual mapping. The key-only module is the worst performing in $T_{en}$ $\leftrightarrow$ $T_{x}$ because it's the first relaxation in the algorithm.  Further analysis of the error cases is in Appendix (\S\ref{apdx:error_analysis}).

\subsection{Information Updation} 
\label{ssec:exp_updation}

\begin{table}[h]
\setlength{\tabcolsep}{1.0pt}
\small
\centering
\begin{tabular}{ l|ccc|ccc} 
\toprule
 \bf Rules & & \multicolumn{1}{c}{\bf Gold} & &  \multicolumn{2}{c}{\bf Predicted}  \\
        & \bf $T_{en}$ $\rightarrow$ $T_{x}$  & \bf $T_{x}$ $\rightarrow$ $T_{y}$    & \bf Live Set & \bf $T_{en}$ $\rightarrow$ $T_{x}$  & \bf $T_{x}$ $\rightarrow$ $T_{y}$   \\
        \midrule
R1     & 20320  & 18055           & 4213 & 21246  & 17675            \\  \hline
R2     & 648    & 502             & 207  & 1395   & 1852             \\ 
R3     & 546    & 399             & 75   & 443    & 347              \\  
R4     & 142    & 151             & 4    & 120    & 147              \\  
R5     & 3507   & 2116            & 784  & 3193   & 1960              \\  
R6     & 5237   & 3047            & 332  & 5062   & 2891             \\ 
R7     & 2748   & 1899            & 990  & 2732   & 1855             \\  
R8     & 25     & 77              & 5    & 29     & 82              \\ \hline
$Al$& 14967 & 9715                & 2851  &    14864 & 10657    \\
\bottomrule
\end{tabular}
\caption{\textbf{Updates on Test Corpora:} Count of the number of updates done by different rules listed in \S \ref{ssec:method_updation}.$Al$ is the number of Alignments. R1-R8 are the rules listed in the same sequential manner as listed in \S \ref{ssec:method_updation}.}
\label{tab:rule_base_update_analyis_test}
\vspace{-1.0em}
\end{table}

\begin{table}[!h]
\small
\centering
\setlength{\tabcolsep}{2.5pt}
\begin{tabular}{ l|ccc} 
\toprule
\bf Type        & \bf Total  & \bf Accept     & \bf Reject    \\
\midrule
Row Transfer     & 461  & 368(79.82\%)           & 93(20.17\%)   \\ 
Value Substitution     & 70   & 52(74.28\%)            & 18(25.72\%)    \\
Append Value    & 72   & 46(63.88\%)            & 26(36.12\%)    \\ 
\hline
Total            & 603  & 466 (77.28\%)           & 136(22.72\%)      \\
\bottomrule
\end{tabular}
\caption{\textbf{Analysis of Human-Assisted Updates:} Accept/Reject rate of different types of edits for human-assisted Wikipedia infobox updates.}
\label{tab:live_updates_results_analysis}
\vspace{-1.0em}
\end{table}

\begin{table}[!h]
\small
\centering
\begin{tabular}{ l|ccc} 
\hline
\bf        Ln Pairs       & \bf Total  & \bf Accept     & \bf Reject    \\
\hline
$T_{en}$ $\rightarrow$ $T_{x}$        & 204  & 161(78.92\%)           & 43(21.07\%)   \\ 
$T_{x}$ $\rightarrow$ $T_{y}$         & 216  & 169(78.25\%)           & 47(21.75\%)    \\
$T_{x}$ $\rightarrow$ $T_{en}$        & 183  & 136(74.31\%)            & 47(25.68\%)    \\ \hline
Total                                 & 603  & 466(77.28\%)           & 137(22.71\%)      \\
\hline
\end{tabular}
\caption{\textbf{Human-Assisted Wikipedia infobox updates:} Accept/Reject rate for different flows of information.}
\label{tab:live_updates_results}
\vspace{-1.5em}
\end{table}

Table \ref{tab:rule_base_update_analyis_test} reports the results of different updation types of rules explained in \S \ref{ssec:method_updation}.
We observe that the row addition rule accounts for the most updated, $\sim$64\%  of total updates for gold and predicted aligned table pairs. The flow of information from high resource to low resource accounts for $\sim$13\% of the remaining updates, whereas a high number of rows too low adds another 8\% of the updates. $\sim$9\% of the updates are done by the value updates rule. All the other rules combined give 8\% of the remaining suggested updates. From the above results, most information gaps can be resolved by row transfer. The magnitude of rules like value updates and multi-key shows that table information needs to be synchronized regularly. Examples of edited infoboxes using the proposed algorithm are shown in Appendix Figures \ref{fig:ProposedUpdates:1} and \ref{fig:ProposedUpdates:2}.

Table \ref{tab:live_updates_results_analysis} reports a similar analysis for human-assisted Wikipedia infobox edits. We also report Wikipedia editors' accept/reject rate for the above-deployed system in Table \ref{tab:live_updates_results}. We obtained an acceptance rate of 77.28\% (as of May 2023), with the highest performance obtained when information flows across non-English languages. The lowest performance is obtained when the information flows from non-English to an English info box. This highlights that our two-step procedure is effective in a real-world scenario. Examples of live updates are shown in Appendix Figures \ref{fig:live_updates:1} and \ref{fig:live_updates:2}. 

\section{Related Works} 
\label{sec:related_works}

\paragraph{Information Alignment.}  
Multilingual Table attribute alignment has been previously addressed via supervised \cite{10.1145/1498759.1498813,zhang2017cross,ta2015model} and unsupervised methods \cite{bouma-etal-2009-cross,nguyen2011multilingual}. Supervised methods trained classifiers on features extracted from multilingual tables. These features include cross-language links, text similarity, and schema features. Unsupervised methods made use of corpus statistics and template/schema matching for alignments.  
Other techniques by \citet{jang2016utilization,nguyen2018automatically} focus on using external knowledge graphs such as DBpedia for the updation of Infoboxes or vice versa. In their experiments, most of these methods use less than three languages, and machine translation is rarely used. Additionally, we don't require manual feature curation for strong supervision. We study the problem more thoroughly with grouped analysis along languages, categories, and keys direction. The works closest to our approach are \citet{nguyen2011multilingual,RINSER2013887}, both of which use cross-language hyperlinks for feature or entity matching. \citet{nguyen2011multilingual} uses translations before calculating text similarity. Utilizing cross-language links can provide a robust alignment supervision signal. In contrast to our approach, we do not use external knowledge or cross-language links for alignments. This additional information is rarely available for languages other than English.


\paragraph{Information Updation.}  
Prior work for information updates \cite{iv-etal-2022-fruit,spangher-etal-2022-newsedits,panthaplackel-etal-2022-updated,Zhang:2020:GCS,Zhang:2020:NED} covers Wikipedia or news articles than semi-structured data like tables. \citet{spangher-etal-2022-newsedits} studies the problem of updating multilingual news articles across different languages over 15 years. They classify the edits as addition, deletion, updates, and retraction. These were the primary intuitions behind our challenge classified  in \S \ref{sec:motivation}. \citet{iv-etal-2022-fruit} focused on automating article updates with new facts using large language models. \citet{panthaplackel-etal-2022-updated} focused on generating updated headlines when presented with new information. Some prior works also focus on the automatic classification of edits on Wikipedia for content moderation and review \cite{sarkar-etal-2019-stre,daxenberger-gurevych-2013-automatically}. Evening modeling editor's behavior for gauging collaborative editing and development of Wikipedia pages has been studied \cite{jaidka-etal-2021-wikitalkedit,yang-etal-2017-identifying-semantic}. Other related works include automated sentence updation based on information arrival \cite{Shah_Schuster_Barzilay_2020,dwivedi2022editeval}. None of these works focus on tables, especially Wikipedia Infoboxes. Also, they fail to address multilingual aspects of information updation.


\section{Conclusion and Future Work} 
\label{sec:conclusion}

Information synchronization is a common issue for semi-structured data across languages.
Taking Wikipedia Infoboxes as our case study, we created \datasetName and proposed a two-step procedure that consists of alignment and updation. The alignment method outperforms baseline approaches with an F1-score greater than 85; the rule-based method received a 77.28 percent approval rate when suggesting updates to Wikipedia.  

We identify the following future directions. \begin{inparaenum}[(a)] 
\item \textit{Beyond Infobox Synchronization.} While our technique is relatively broad, it is optimized for Wikipedia Infoboxes. We want to test whether the strategy applies to technical, scientific, legal, and medical domain tables \cite{wang-etal-2013-transfer,10.1145/3004296}. It will also be intriguing to widen the updating rules to include social, economic, and cultural aspects.
\item \textit{Beyond Pairwise Alignment.} Currently, independent language pairs are considered for (bi) alignment. However, multiple languages can be utilized jointly for (multi) alignment. 
\item \textit{Beyond Pairwise Updates.} Similar to (multi) alignment, one can jointly update all language variants simultaneously. 
This can be done in two ways: (1.) \textit{With English as pivot language :} To update across all languages. Here, English act as a central server with message passing. (2.) \textit{Round-Robin Fashion:} where pairwise language updates between language pairs are transferred in a round-robin ring across all language pairs. In every update, we selected a leader similar to a leader election in distributed systems.
\item \textit{Joint Alignment and Updation.} Even while our current approach is accurate, it employs a two-step process for synchronization, namely alignment followed by updating. We want to create rapid approaches aligning and updating in a single step.
\item \textit{Text for Updation}: Our method doesn't consider Wikipedia articles for updating tables \cite{10.1145/1871437.1871698,10.1145/3184558.3191647,10.1145/2396761.2398627}.
\end{inparaenum}
\section*{Limitations}
\label{sec:limitations}
We only consider 14 languages and 21 categories, whereas Wikipedia has pages in more than 300 languages and 200 broad categories. Increasing the scale and diversity will further improve method generalization. Our proposed method relies on the good multilingual translation of key and value from table pairs. Although we use key, value, and category together for better context, enhancement in table translation \cite{minhas-etal-2022-xinfotabs} will benefit our approach. Because our rule-based system requires manual intervention, it has automation limits. Upgrading to completely automated methods based on a large language model may be advantageous. We are only considering updates for semi-structured tables. However, updating other page elements, such as images and article text, could also be considered. Although a direct expansion of our method to a multi-modal setting is complex \cite{suzuki2012good}.  

\section*{Ethics Statement}
\label{sec:ethics}
We aimed to create a balanced, bias-free dataset regarding demographic and socioeconomic factors. We picked a wide range of languages, even those with limited resources, and we also ensured that the categories were diversified. Humans curate the majority of information on Wikipedia. Using unrestricted automated tools for edits might result in biased information. For this reason, we adhere to the "human in the loop" methodology \cite{smith2020keeping} for editing Wikipedia. Additionally, we follow Wikipedia editing guidelines\footnote{\url{https://en.wikipedia.org/wiki/Wikipedia:List_of_policies_and_guidelines}}, rule set\footnote{\url{https://en.wikipedia.org/wiki/Wikipedia:Simplified_ruleset}}, and policies\footnote{\url{https://en.m.wikipedia.org/wiki/Wikipedia:Editing_policy}} for all manual edits. Therefore, we ask the community to use our method only as a recommendation tool for revising Wikipedia. As a result, we ask that the community utilize \datasetName strictly for scientific and non-commercial purposes from this point forward.

\section*{Acknowledgements}
We thank members of the Utah NLP group for their valuable insights and suggestions at various stages of the project, and reviewers for their helpful comments. Additionally, we appreciate the inputs provided by Vivek Srikumar and Ellen Riloff. Vivek Gupta acknowledges support from Bloomberg’s Data Science Ph.D. Fellowship.

\bibliography{custom}
\bibliographystyle{acl_natbib}

\appendix
\section{Appendix}
\label{sec:appendix}


\subsection{Table Extraction Details}
\label{apdx:ted}

\begin{table}[h]
\setlength{\tabcolsep}{1.8pt}
\centering

\small
\begin{tabular}{ l|cc|l|cc } 

\toprule

\bf Category    & \bf Entities   &  $std\_dev$   & \bf Category & \bf Entities   &  $std\_dev$  \\ 
\midrule
 Airport             & 2563       & 5.03     &   Country             & 259        & 10.28       \\ 
 Album               & 840        & 3.81      &  Diseases            & 462        & 4.20        \\ 
 Animal              & 368        & 3.37       &  Food                & 692        & 4.34        \\  
 Athlete             & 369        & 5.80        & Medicine            & 334        & 9.58        \\ 
 Book                & 218        & 5.24        & Monument            & 203        & 5.23        \\  
 City                & 262        & 7.95        & Movie               & 1524       & 6.75        \\ 
 College             & 202        & 5.83        & Musician            & 284        & 5.09        \\  
 Company             & 267        & 6.87        & Nobel               & 967        & 5.29        \\ 
 Painting            & 743        & 3.51        &  Stadium             & 742        & 5.86        \\ 
 Person              & 198        & 6.32        &  Shows               & 1044       & 6.83        \\ 
 Planet              & 188        & 8.46     & & &   \\

\hline
\end{tabular}
\caption{\textbf{Missing information Analysis in Categories}:- For each category unique number of entities and their average standard deviation  across languages.}
\label{tab:apdx_infosync_category_wise_entities}
\end{table}

\begin{table}[h]
\small
\centering
\setlength{\tabcolsep}{5.5pt}
\begin{tabular}{ l|c |l|c } 
\toprule
\bf C1     & \bf  Row Diff  & \bf C1     & \bf Row Diff \\
\midrule
 af             & 5.28   & hi             & 5.06            \\ 
 ar             & 5.84          
 &  ko             & 4.30          \\
 ceb            & 3.33           
 &  nl             & 3.86          \\ 
 de             & 5.96           
 &  ru             & 4.1          \\ 
 en             & 4.80          
 &  sv             & 3.92          \\ 
 es             & 5.17           
 &  tr             & 4.23          \\ 
 fr             & 4.42           
& zh             & 4.76          \\ 
\bottomrule
\end{tabular}
\caption{\textbf{Row Difference Across Paired Languages}:- Column 2 shows average row count difference between languages for all entities.}
\label{tab:apdx_infosync_row_pair_transfer}

\end{table}


Table formats and HTML code styles differ from one language to another and even across categories in the same language. Extraction is modified to handle these variations, which requires the following steps: \begin{inparaenum}[(a)] \item \emph{Detecting Infoboxes:} We locate Wikipedia infoboxes that appear in at least five languages. \item \emph{Extracting HTML:} After detection, we extract HTML and preprocess to remove images, links, and signatures. \item \emph{Table Representation:} we convert the extracted table and store them in JSON. \end{inparaenum}

\textit{Row Difference Across Paired Languages:} There is substantial variation in the number of rows for infobox across different languages, i.e., rows difference = $\frac{1}{|L|}$$\sum_{ln \in L\setminus{C1}}  ||R_{c1}| - |R_{ln}||$, where $L$ is set of all 14 languages under consideration. Table \ref{tab:apdx_infosync_row_pair_transfer} shows that German followed by Arabic and Afrikaans, has the highest row difference. This indicates that tables in these languages are incomplete (with missing rows). 

%

\subsection{Table Updation Examples}
\label{apdx:iur} 
\begin{figure*}[t]
       \centering
	\includegraphics[scale=0.4]{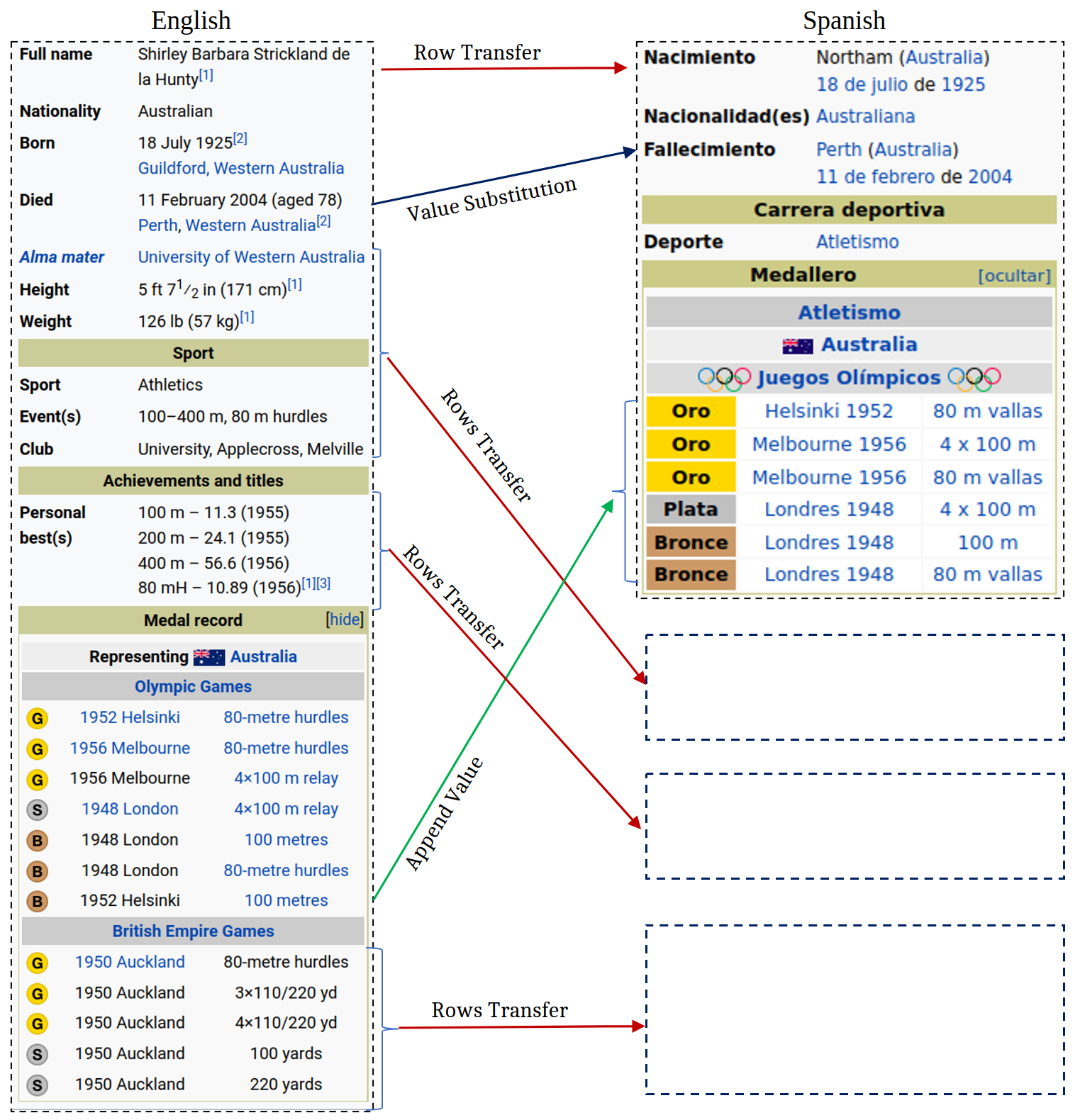}
    \caption{Update Example:- "Shirley Strickland de la Hunty " Infoboxes for two languages, i.e., English and Spanish. Shows rows transfer for missing information. Value substitution because "Aged 78" is absent in Died. One medal information (Bronze,1952, 100m) is added to the medal tally.} 
\label{figure::apdx:Update} 
\end{figure*}

An example of table updation is shown in the Figure \ref{figure::apdx:Update}.

\subsection{Precision and Recall}
\label{apdx:prec_recall}
We also evaluated precision-recall values in information alignment for matched and unmatched scores (\S \ref{ssec:exp_alignment}). Precision recall values for $T_{en}$ $\leftrightarrow$ $T_{x}$, $T_{x}$ $\leftrightarrow$ $T_{y}$, $T_{en}$ $\xleftrightarrow{*}$ $T_{hi}$ and $T_{en}$ $\xleftrightarrow{*}$ $T_{zh}$ are reported in Tables \ref{tab:appendix_eng_x_alignment}, \ref{tab:appendix_x_y_alignment}, \ref{tab:appendix_en_hi_alignment}, and \ref{tab:appendix_en_zh_alignment}, respectively.

\subsection{Algorithm Coverage} \label{apdx:coverage}
\begin{table*}[h]
\small
\centering
\setlength{\tabcolsep}{6.5pt}
\begin{tabular}{ l|ccccc| ccccc } 
\toprule
\bf Ablation & & \multicolumn{3}{c}{\bf  $T_{en}$ $\leftrightarrow$ $T_{x}$} & &  &\multicolumn{3}{c}{\bf  $T_{x}$ $\leftrightarrow$ $T_{y}$} & \\ 
 &  Corpus & Key &  K-V-Bi	 &  K-V-Uni	&  Multi &  Corpus   & Key    &   K-V-Bi	 &  K-V-Uni	&  Multi \\
\midrule
w/o        & 16.28  & 40.83  &  58.17 & 71.95 &\bf 72.54  &  17.15    & 39.78&  57.53 & 67.58 &\bf  67.96 \\ \hline
Corpus       &  -     & 33.15  &  50.17 & 74.69 &\bf  75.3  & -     &   25.04  & 49.82 & 64.98 &\bf  65.41 \\ 
Key	       &  16.28 & -      &  57.88 & 71.14 &\bf  71.6  & 17.15 &     -    & 55.05 &  64.8 &\bf  65.1   \\
K-V-Bi	   &  16.28 & 38.88  &  -     & 71.9  &\bf  72.3  & 17.15 &   37.83  &    - & 70.32 &\bf  70.58  \\
K-V-Uni	   &  16.28 & 46.19  &  21.34 & -     &\bf  67.53 & 17.15 &   40.19  & 62.91 & -    &\bf  63.59  \\
Multi	   &  16.28 & 40.96  &  58.4  &\bf 72.23 &     -  & 17.15 &   36.36  & 55.03 &\bf 67.13 &  -    \\
\bottomrule
\end{tabular}
\caption{\textbf{Coverage Ablation:}  $T_{en}$ $\leftrightarrow$ $T_{x}$ and  $T_{x}$ $\leftrightarrow$ $T_{y}$.}
\label{tab:apdx_coverage_ablations}
\end{table*}

We measure the coverage on the entire corpus, the rate of rows aligned w.r.t. the smaller table in a table pair. 
Table \ref{tab:apdx_coverage_ablations} reports ablations results of coverage for various modules.
Our proposed method aligns 72.54\% and 67.96\% of rows for $T_{en}$ $\leftrightarrow$ $T_{x}$ and $T_{x}$ $\leftrightarrow$ $T_{y}$, respectively.  Corpus-based is the most constrained module, focusing more on precision; hence removing corpus-based gives better coverage for both cases. Key-Only-Unidirectional is the most important module for coverage, followed by the Key-Only module for both cases. 

\subsection{Domain and Language Wise Analysis}
\label{apdx:domain_language_wise}

Table \ref{tab:apdx_alignment_language_wise}, \ref{tab:apdx_alignment_domain_wise}, and \ref{tab:apdx_alignment_key_wise} show the performance of our proposed method grouped by languages, domains, and keys, respectively.
\begin{table}[h]
\small
\setlength{\tabcolsep}{1.5pt}
\centering
\begin{tabular}{ c|cc |c|cc } 
\hline
$T_{en}$ $\leftrightarrow$ $T_{x}$         & \bf Match    & \bf UnMatch     & $T_{x}$ $\leftrightarrow$ $T_{y}$             & \bf Match    & \bf UnMatch       \\
\hline
 af               & 88.08       & 89.48    & de $\leftrightarrow$ hi               & 88.85       & 90.4  \\ 
ar               & 85.24       & 88.77    & de $\leftrightarrow$ ko               & 85.27       & 88.7      \\  
ceb              & 85.17       & 91.07    & fr $\leftrightarrow$ ar               & 85.35       & 87.21     \\  
de               & 85.41       & 86.65    & fr $\leftrightarrow$ de               & 84.97       & 88.94     \\ 
es               & 89.83       & 89.7     & fr $\leftrightarrow$ hi               & 83.95       & 84.58      \\  
fr               & 89.41       & 89.8     & fr $\leftrightarrow$ ko               & 83.59       & 84.36     \\ 
hi               & 90.56       & 87.07    & fr $\leftrightarrow$ ru               & 87.63       & 88.83     \\ 
ko               & 85.69       & 86.22    & hi $\leftrightarrow$ ar               & 84.33       & 89.38     \\  
nl               & 86.4        & 90.28    & ko $\leftrightarrow$ ar               & 82.18       & 89.08     \\  
ru               & 87.46       & 88.54    & ko $\leftrightarrow$ hi               & 78.8        & 83.03     \\ 
sv               & 84.89       & 86.76    & ru $\leftrightarrow$ ar               & 82.18       & 86.96     \\  
tr               & 92.07       & 91.3     & ru $\leftrightarrow$ de               & 89.93       & 91.92     \\  
zh               & 91.61       & 89.31    & ru $\leftrightarrow$ hi               & 82.38       & 87.78      \\ 
				& 				& 		  & ru $\leftrightarrow$ ko               & 81.62       & 84.47     \\ 
				& 				&		 & 	de $\leftrightarrow$ ar               & 78.05       & 87.23  \\
 
\hline
\end{tabular}
\caption{\textbf{Language Wise Analysis} :-Alignment F1-score reported for same language for  $T_{en}$ $\leftrightarrow$ $T_{x}$ and $T_{x}$ $\leftrightarrow$ $T_{y}$ averaged over all entities.  }
\label{tab:apdx_alignment_language_wise}
\end{table}

\begin{table}[h]
\small
\setlength{\tabcolsep}{2.5pt}
\centering
\begin{tabular}{ l|cc|cc } 

\toprule
 & \multicolumn{2}{c}{\bf $T_{en}$ $\leftrightarrow$ $T_{x}$ } & \multicolumn{2}{|c}{\bf $T_{x}$ $\leftrightarrow$ $T_{y}$ }  \\

\bf Category    &     \bf Match    & \bf UnMatch & \bf Match    & \bf UnMatch       \\
\midrule
 Airport             & 79.77       & 82.64   & 85.79       & 90.9    \\ 
 Album               & 93.9        & 91.33   & 88.6        & 85.01  \\ 
 Animal              & 93.79       & 94.2    & 90          & 96.24 \\ 
 Athlete             & 86.6        & 90.21   & 83.75       & 88.81  \\ 
 Book                & 86.48       & 90.96   & 81.29       & 83.13   \\ 
 City                & 86.14       & 93.67   & 77.4        & 86.6   \\ 
 College             & 82.47       & 87.53   & 81.05       & 86.24  \\ 
 Company             & 87.49       & 85.15   & 85.5        & 86.7  \\ 
 Country             & 86.38       & 92.47   & 86.53       & 92.32  \\ 
 Food                & 88.58       & 90.04   & 85.65       & 91.67  \\ 
 Monument            & 84.86       & 86.14   & 87.66       & 89.6  \\ 
 Movie               & 91.2        & 85.7    & 74.33       & 76.19  \\ 
 Musician            & 89.47       & 85.62   & 89.04       & 93.27  \\ 
 Nobel               & 88.2        & 91.08   & 88.84       & 87.1  \\ 
 Painting            & 90.27       & 82.35   & 86.52       & 89.72  \\ 
 Person              & 87.37       & 87.79   & 79.85       & 87.74  \\ 
 Planet              & 90.93       & 85.77   & 85.01       & 87.18  \\ 
 Shows               & 91.23       & 88.89   & 83.65       & 78.84 \\ 
 Stadium             & 88.59       & 87.72   & 83.2        & 77.38  \\ 
\hline
\end{tabular}
\caption{\textbf{Category Wise Analysis} :- Alignment F1-score reported for same group entities average over all languages.}
\label{tab:apdx_alignment_domain_wise}
\end{table}
    
\begin{table}[h]
\small
\centering
\setlength{\tabcolsep}{1.5pt}
\begin{tabular}{ l|c|cc } 
\hline
\bf Key Freq     & \bf Range           & \bf \# of Keys (all)    & \bf Avg Score     \\
\hline
 High    & $100 \le x$     & 33          & 90.71  \\ 
 Mid    & $50\le x\le100$  & 49           & 89.33  \\
 Low    & $x\le 50$       & 700           & 81.82   \\
\hline
\end{tabular}
\caption{\textbf{Key Wise Analysis}:- F1-Score report for grouped keys.}
\label{tab:apdx_alignment_key_wise}
\end{table}

\textit{Group-wise Analysis.} 
From Table \ref{tab:apdx_alignment_language_wise}, for $T_{en}$ $\leftrightarrow$ $T_{x}$, Cebuano, Arabic, German, and Dutch are the worst performing languages with F1-score close to 85 for alignment. Whereas Turkish, Chinese, and Hindi have F1-score greater than 90. Korean, German, and Swedish are the lowest-performing language groups, with an F1-Score close to 86 for unaligned settings. Cebuano, Turkish, and Dutch get the highest score for unaligned metrics (greater than 90). For non-English language pairs, the lowest F1-score for match table pairs is observed for German-Arabic and Hindi-Korean pairs with an F1-score close to 78, as shown in Table \ref{tab:apdx_alignment_language_wise}. The highest F1-score is observed for Russian-German and Hindi-German, with F1-scores exceeding 88.8. For unmatched data, Korean-Hindi, French-Hindi, French-Korean, and Russian-Korean pairs have the lowest F1 scores, less than 85. In contrast, German-Hindi and Russion-German have exceeded the unaligned F1-Score of 90. 

\textit{Category-wise Analysis.} 
As reported in Table \ref{tab:apdx_alignment_domain_wise}, our method performs worst in Airport and College categories for match settings when one of the languages is English. For non-English match settings, Movie and City are the worst-performing categories. For unmatch setting with English as one of the languages, Airport and Painting have the lowest F1-score, whereas Movie and Stadium have the most inferior performance for non-English languages. 

\textit{Key-wise Analysis.} Table \ref{tab:apdx_alignment_key_wise} shows the average F1-scores across tables for frequent and non-frequent keys. We observed an F1-score degradation of 10 points for rare keys with a low occurrence compared to frequent keys.

\subsection{Ablation Study} \label{apx:ablation_study}

\begin{table*}[h]
\small
\centering
\setlength{\tabcolsep}{3.0pt}
\begin{tabular}{ l|cccc | cccc } 
\toprule
\bf  Ablation & & \multicolumn{2}{c}{\bf Match} & &  &\multicolumn{2}{c}{\bf UnMatch} & \\ 
 & \bf  $T_{en}$ $\leftrightarrow$ $T_{x}$ & \bf  $T_{x}$ $\leftrightarrow$ $T_{y}$ & \bf $T_{en}$ $\xleftrightarrow{*}$ $T_{hi}$ & \bf  $T_{en}$ $\xleftrightarrow{*}$ $T_{zh}$ & \bf  $T_{en}$ $\leftrightarrow$ $T_{x}$ & \bf  $T_{x}$ $\leftrightarrow$ $T_{y}$ & \bf  $T_{en}$ $\xleftrightarrow{*}$ $T_{hi}$ & \bf  $T_{en}$ $\xleftrightarrow{*}$ $T_{zh}$ \\
\midrule
 Corpus-based  & 86.67 & 82.3  & 89.13 & 92.33 & 87.95 & 87.03 & 83.11 & 87.38 \\ 
 Key Only    & 89    & 80.09 & 87.35 & 91.49 & 89.42 & 85.88 & 79.83 & 87.13 \\
 Key-Val-Bi  & 84.98 & 75.39 & 86.95 & 90.41 & 86.39 & 82.06 & 80.48 & 84.4\\
 Key-Val-Uni & 87.73 & 79.35 & 90    & 92.67 & 89.03 & 85.35 & 84.83 & 88.74  \\
 Multi-Key	& 87.89 & 84.33 & -     & -     & 89.52 & 85.42 & -     & -     \\
 \hline
 w/o         & 87.91 & 84.36 & 90.14 & 92.8  & 89.03 & 85.46 & 84.98 & 88.17  \\
\bottomrule
\end{tabular}
\caption{\textbf{Ablation Study of Matched and UnMatch Score :} i.e. F1-Score for all test sets of \datasetName.}
\label{tab:apdx_aligned_unaligned_ablation}
\end{table*}

We report ablation performance to highlight the significance of each module in Table \ref{tab:apdx_aligned_unaligned_ablation}. Key-Value-Bidirectional mapping (two-way) is the most critical module, followed by Key Only corpus-based modules. We also observe Uni-directional mapping being the second most important for non-English alignments. The multi-key module was consistently was least significant for the same reason as the discussion above (very few instances). Similar observations were valid for unmatching scores.

\subsection{Further Details: Error Analysis}
\label{apdx:error_analysis}

We discussed challenges to table information synchronization across languages in \S \ref{sec:motivation}. Table \ref{tab:aligned_error_anlysis} (main paper) shows the number of instances of these challenges in evaluation for matched cases after applying various modules of the alignment algorithm. \label{apdx:ted:ea}
\begin{itemize}
\item Corpus-based module solves approximately (40\%, 56\%)  of outdated information,(31\%, 21\%) of schema variation, (10\%, 30\%) of language variation, (13\%, 25\%) of unnormalized information and (37\%, 26\%) of erroneous entity linking challenges in  $T_{en}$ $\leftrightarrow$ $T_{x}$ and $T_{x}$ $\leftrightarrow$ $T_{y}$, respectively.
\item Further adding of key only similarity module resolves extra (24\%, 13\%) of outdated information, (36\%, 21\%) of schema variation, (13\%, 5\%) of language variation, (19\%, 17\%) of unnormalized information and (22\%, 8\%) of erroneous entity linking challenges in  $T_{en}$ $\leftrightarrow$ $T_{x}$ and $T_{x}$ $\leftrightarrow$ $T_{y}$, respectively.
\item Applying key-value-bidirectional module resolves another (12\%, 13.5\%) of outdated information, (18\%, 6.6\%) of information representation, {54\%, 45\%} of language variation, (40\%, 46\%) of unnormalized information and (34\%, 53\%) of erroneous entity linking challenges in  $T_{en}$ $\leftrightarrow$ $T_{x}$ and $T_{x}$ $\leftrightarrow$ $T_{y}$, respectively.
\item Key-Val-Unidirectional and Multi-key together solves another (18.5\%, 7.5\%) of the information representation in  $T_{en}$ $\leftrightarrow$ $T_{x}$ and $T_{x}$ $\leftrightarrow$ $T_{y}$, respectively, but are not effective against other challenges.
\end{itemize}

\begin{table*}[h]
\small
\centering
\begin{tabular}{l|ccc|ccc} 
\toprule
&  &\multicolumn{1}{c}{\bf Match}& & &\multicolumn{1}{c}{\bf UnMatch} &\\
\bf Alignment &\bf Precision &\bf Recall &\bf F1 &\bf Precision &\bf Recall &\bf F1  \\
\midrule
 Corpus-based     & 93.51 & 46.22 & 61.86  & 55.66    & 96.17 & 70.51 \\ 
+ Key Only	    & 88.09	&58.62	&70.4	 & 60.75	& 94.16	& 73.85 \\
+ Key-Value-Bi	& 89.6	&85.89	&87.71   & 85.87	& 93.47	& 89.51 \\
+ Key-Value-Uni	& 89.3 &86.52	&87.89	 & 86.24    & 93.07	& 89.52	\\
+ Multi-Key	    & 88.85	&86.99	&87.91	 & 86.51	& 92.27	& 89.3	\\
\hline
\end{tabular}
\caption{$T_{en}$ $\leftrightarrow$ $T_{x}$ alignment performance on Human-Annotated Test Data}
\label{tab:appendix_eng_x_alignment}
\end{table*}

\begin{table*}[h]
\small
\centering
\begin{tabular}{ l|ccc|ccc } 
\toprule
&  &\multicolumn{1}{c}{\bf Match}& & &\multicolumn{1}{c}{\bf UnMatch} &\\
\bf Alignment &\bf Precision &\bf Recall &\bf F1 &\bf Precision &\bf Recall &\bf F1  \\
\midrule
 Corpus-based     & 75.68 & 45.38 & 56.74 & 58.9  & 91.71 & 71.73  \\ 
+ Key Only	    & 74.45	& 58.62	& 62.14	& 62.44	& 89.37	& 73.52	 \\
+ Key-Value-Bi  & 82.78	& 85.66	& 84.2  & 82.53	& 88.73	& 85.52\\
+ Key-Value-Uni	& 82.2  & 86.58	& 84.33	& 82.94 & 88.05	& 85.42 \\
+ Multi-Key	    & 82.16	& 86.68	& 84.36	& 83.05	& 88.01	& 85.46	 \\
\bottomrule
\end{tabular}
\caption{$T_{x}$ $\leftrightarrow$ $T_{y}$ alignment performance on Human-Annotated Test Data.}
\label{tab:appendix_x_y_alignment}
\end{table*}

\begin{table*}[h]
\small
\centering
\begin{tabular}{ l|ccc|ccc} 
\toprule
&  &\multicolumn{1}{c}{\bf Match}& & &\multicolumn{1}{c}{\bf UnMatch} &\\
\bf Alignment &\bf Precision &\bf Recall &\bf F1 &\bf Precision &\bf Recall &\bf F1  \\
\midrule
 Corpus-based     & 94.81 & 41.1  & 57.34 & 37.55 & 96.19 & 71.73\\ 
+ Key Only	    & 92.04	& 61.04	& 73.4	& 46.6	& 94.81	& 73.52 \\
+ Key-Value-Bi	& 87.65	& 85.89	& 90.07 & 77.37	& 88.73	& 85.52\\
+ Key-Value-Uni	& 88.59 & 86.52	& 90.34	& 78.53 & 88.05	& 85.42\\
+ Multi-Key	    & 91.15	& 88.59	& 90.14 & 78.52	& 88.01	& 85.46		 \\
\bottomrule
\end{tabular}
\caption{$T_{en}$ $\xleftrightarrow{*}$ $T_{hi}$ alignment performance on Human-Annotated Test Data.}
\label{tab:appendix_en_hi_alignment}
\end{table*}

\begin{table*}[h]
\small
\centering
\begin{tabular}{  l|ccc|ccc } 
\toprule
&  &\multicolumn{1}{c}{\bf Match}& & &\multicolumn{1}{c}{\bf UnMatch} &\\
\bf Alignment &\bf Precision &\bf Recall &\bf F1 &\bf Precision &\bf Recall &\bf F1  \\
\midrule
 Corpus-based     & 89.94 & 56.41 & 69.33 & 47.19 & 95.26 & 63.11 \\ 
+ Key Only	    & 88.78	& 64.43	& 74.67	& 51.74 & 91.99 & 66.23 \\
+ Key-Value-Bi	& 92.38	& 93.7	& 93.04 & 86.73 & 91.81 & 89.2  \\
+ Key-Value-Uni	& 92.13 & 94.13	& 93.12	& 86.75 & 90.58 & 88.62 \\
+ Multi-Key	    & 91.51	& 94.13	& 92.8	& 86.73 & 89.66 & 88.17 \\
\bottomrule
\end{tabular}
\caption{$T_{en}$ $\xleftrightarrow{*}$ $T_{zh}$ alignment performance on Human-Annotated Test Data.}
\label{tab:appendix_en_zh_alignment}
\end{table*}

\subsection{Other Related Work} 
\label{apdx:other_related_works}
\paragraph{Tabular Reasoning.} 

Addressing NLP tasks on semi-structured tabular data has received substantial attention. There is work on tabular NLI \citep{gupta-etal-2020-infotabs, chen2019tabfact,gupta-etal-2022-right}, question-answering task \citep[ and others]{zhang2020tablesurvey,zhu-etal-2021-tat,pasupat:15,Abbas2016WikiQAA,sun2016table,chen2021kace,chen2020hybridqa,lin2020bridging,zayats2021representations, oguz2020unified} and table-to-text generation \citep{zhang2020tablesum,parikh2020totto, radev2020dart,yoran2021turning,chen2020open}. 

\paragraph{Tabular Representation and Learning.} There are also several works representing Wikipedia tables, such papers are TAPAS \citep{herzig-etal-2020-tapas}, StrucBERT \citep{trabelsi2022structbert}, Table2vec \citep{zhang2020table2vec}, TaBERT \citep{yin-etal-2020-tabert}, TABBIE \citep{iida-etal-2021-tabbie}, TabStruc \citep{zhang-etal-2020-table}, TabGCN \citep{pramanick2021joint}, RCI \citep{glass-etal-2021-capturing}, TURL \citep{10.1145/3542700.3542709}, and TableFormer \citep{yang2022tableformer}. Some papers such as \citep[ and others]{yu2018spider,yu2020grappa,eisenschlos:20, neeraja-etal-2021-infotabskg, Mller2021TAPASAS, somepalli2021saint,shankarampeta-etal-2022-enhancing,dong2022table} study pre-training for tabular tasks.  Paper related to tabular probing includes \citep{koleva2022analysis,gupta-etal-2022-model}.

\paragraph{Tabular Datasets.} There are several tabular task datasets on (a.) tabular NLI: \citep[ and others]{gupta-etal-2020-infotabs,Rozen2019DiversifyYD,Mller2021TAPASAS,Kaushik2020Learning,Xiong2020Pretrained,chen2019tabfact,eisenschlos:20,chen-etal-2020-logic2text}; (b.) Tabular QA: WikiTableQA \cite{pasupat2015compositional}, HybridQA \cite{chen2020hybridqa, zayats2021representations, oguz2020unified},WikiSQL \cite{iyyer-etal-2017-search}, SQUALL \cite{ferre2012squall, shi-etal-2020-potential}, OpenTableQA \cite{chen2020open}, FinQA \cite{chen-etal-2021-finqa}, FeTaQA \cite{nan-etal-2022-fetaqa}, TAT-QA \cite{zhu-etal-2021-tat}, SQA \cite{iyyer-etal-2017-search}, NQ-Tables \cite{herzig-etal-2021-open}; (c.) and Table Generation: ToTTo \cite{parikh2020totto}, Turing Tables \cite{yoran2021turning}, LogicNLG \cite{chen-etal-2020-logic2text}. 

Furthermore, there are also several works discussed on web table extraction, retrieval, and augmentation \cite{zhang2020web}, and utilizing the transformers model for table representation \cite{10.14778/3554821.3554890}.

\begin{figure*}
    \centering
    \includegraphics[scale=0.6]{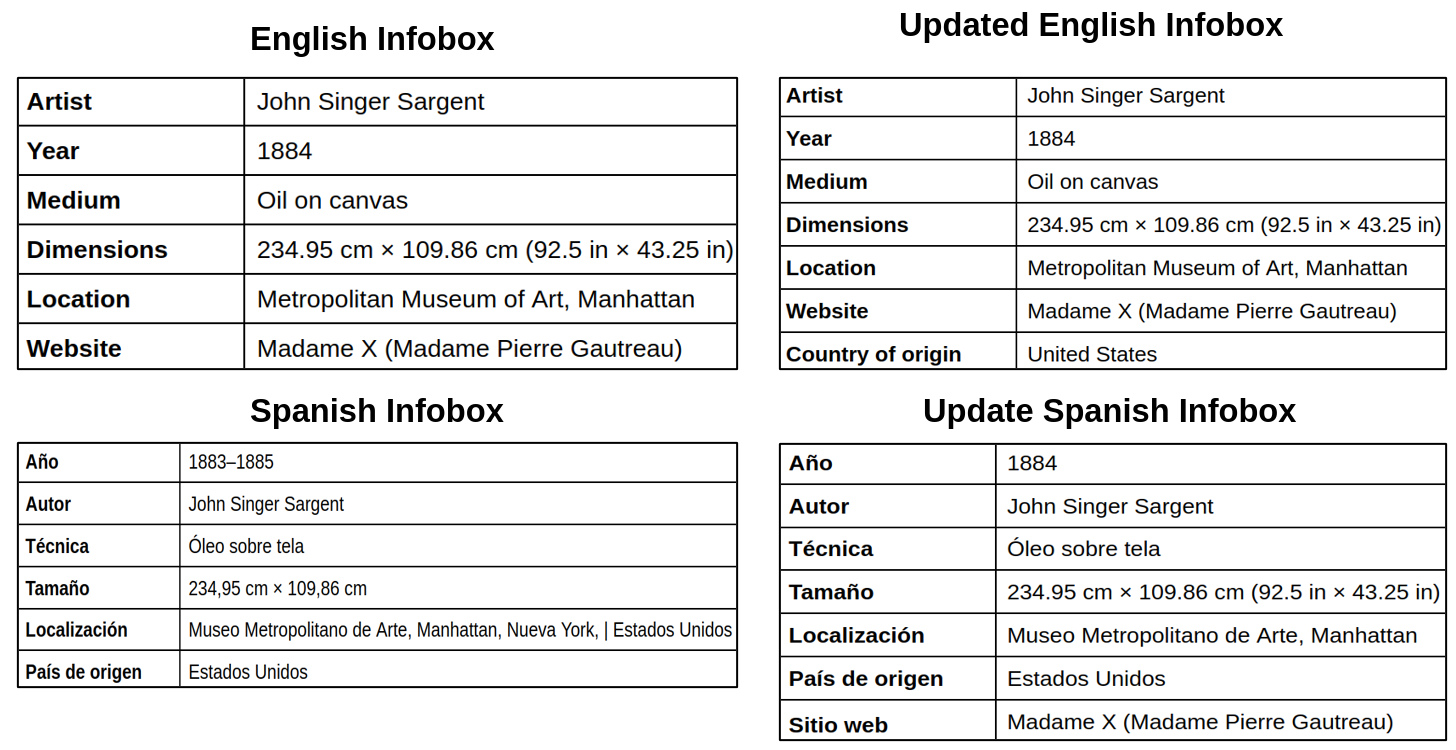}
    \caption{\textbf{Example From Update Algorithm Proposed}: Update English Infobox is obtained by using Spanish Infobox as a reference and vice versa."Country of origin" is updated in English infobox and "website" is updated in Spanish infobox.}
    \label{fig:ProposedUpdates:1}
\end{figure*}

\begin{figure*}
    \centering
    \includegraphics[scale=0.55]{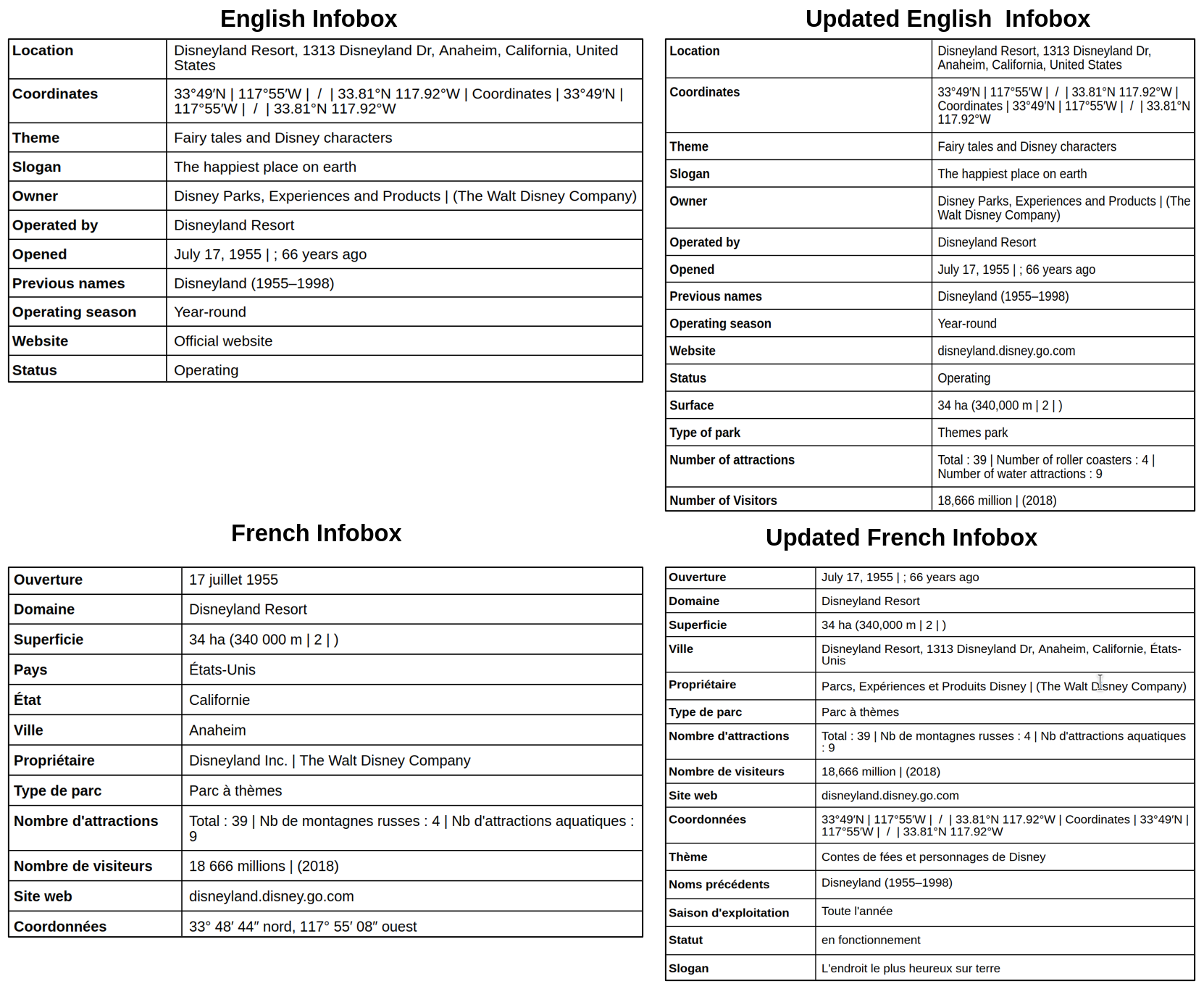}
    \caption{\textbf{Example From Update Algorithm Proposed}: Update English Infobox is obtained by using French Infobox as a reference and vice versa. Multiple keys are updated in both infoboxes "Opened," "Location," "Owner," "Coordinates," "Operating Season," and "Slogan" in French, and "number of visitors," "surface," "Type of park," number of attractions" are updated in English infobox.}
    \label{fig:ProposedUpdates:2}
\end{figure*}

\begin{figure*}
    \centering
    \includegraphics[scale=0.42]{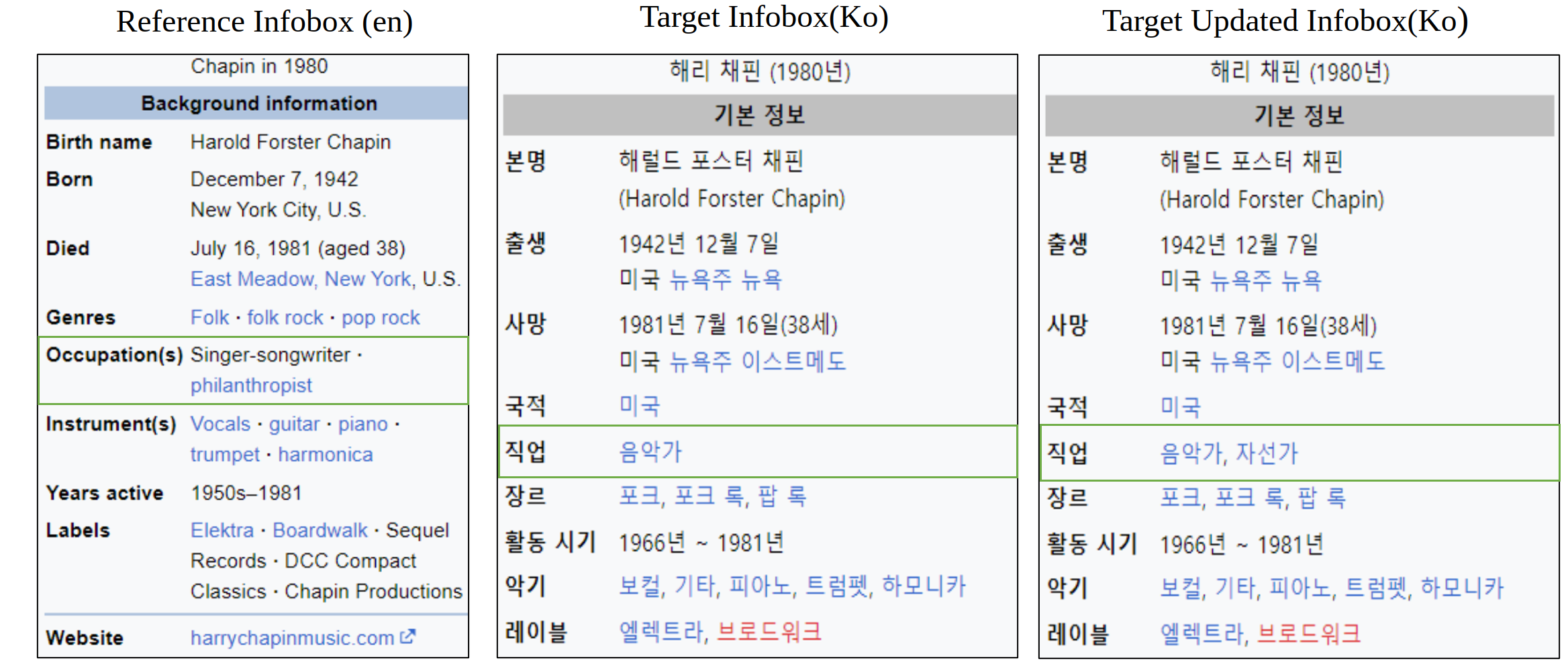}
    \caption{\textbf{Example From Live Updates}: In the above figure, the Target infobox needs to be updated using Reference infobox(available in English version) as extra/grounding information. The updated infobox is shown in column 3, where the key 'job' is updated. This is an example of "Value substitution," as in Table 8. The red box highlights the updated information. }
    \label{fig:live_updates:1}
\end{figure*}

\begin{figure*}
    \centering
    \includegraphics[scale=0.5]{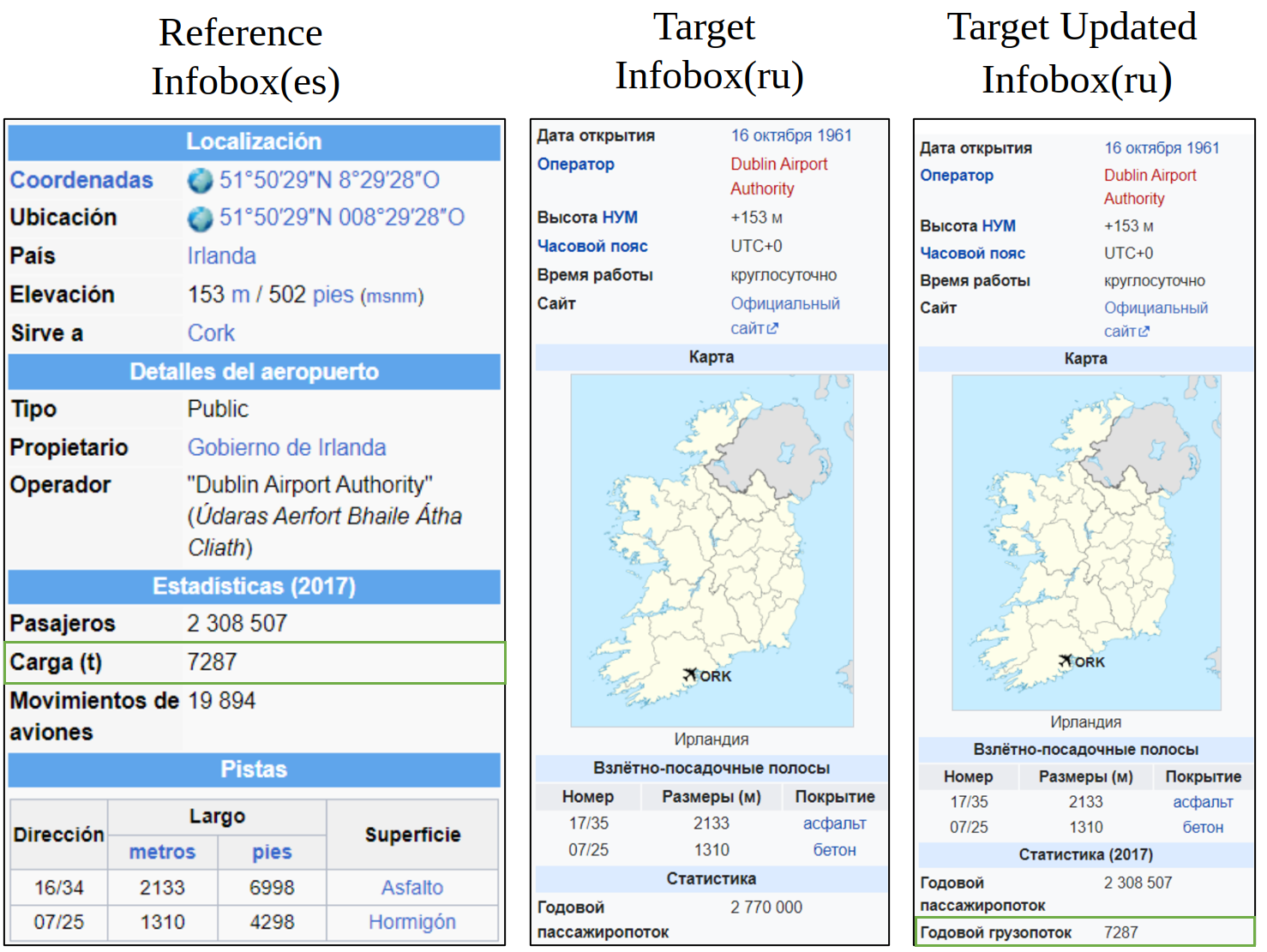}
    \caption{\textbf{Example From Live Updates}: In the above figure, the target infobox needs to be updated using a reference infobox as extra/grounding information. The updated Infobox is shown in column 3, where the 'Load/Cargo Traffic' key is updated. This is an example of Row Addition, as referred to in Table 8. The red box highlights the updated information. }
    \label{fig:live_updates:2}
\end{figure*}

\end{document}